\documentclass{article}
\usepackage[final]{colm2025_conference}
\usepackage{microtype}
\usepackage{hyperref}
\usepackage{url}
\usepackage{booktabs}
\usepackage{graphicx}
\usepackage{multirow}
\usepackage[T1]{fontenc}
\usepackage{xspace}
\usepackage{fontawesome}
\usepackage{float}
\usepackage{placeins}

\definecolor{darkblue}{rgb}{0, 0, 0.5}
\hypersetup{colorlinks=true, citecolor=darkblue, linkcolor=darkblue, urlcolor=darkblue}

\usepackage{pifont}

\newcommand{\docblocks}{\textsc{DocBlocks}\xspace}

\title{Multilingual Contextualization of Large Language Models for Document-Level Machine Translation}

\author{
\textbf{Miguel Moura Ramos}$^{1,2}$ \quad
\textbf{Patrick Fernandes}$^{1,2,3}$ \quad
\textbf{Sweta Agrawal}$^{2}$ \quad \\
\textbf{André F. T. Martins}$^{1,2,4}$ \quad    
\\
$^1$Instituto Superior Técnico, Universidade de Lisboa (ELLIS Unit Lisbon)
\\
$^2$Instituto de Telecomunicações\quad 
$^3$Carnegie Mellon University\quad
$^4$Unbabel \\
}

\begin{document}

\maketitle

\begin{abstract}
Large language models (LLMs) have demonstrated strong performance in sentence-level machine translation, but scaling to document-level translation remains challenging, particularly in modeling long-range dependencies and discourse phenomena across sentences and paragraphs.
In this work, we propose a method to improve LLM-based long-document translation through targeted fine-tuning on high-quality document-level data, which we curate and introduce as \docblocks.
Our approach supports multiple translation paradigms, including direct document-to-document and chunk-level translation, by integrating instructions both with and without surrounding context. This enables models to better capture cross-sentence dependencies while maintaining strong sentence-level translation performance.
Experimental results show that incorporating multiple translation paradigms improves document-level translation quality and inference speed compared to prompting and agent-based methods.
\end{abstract}

\section{Introduction}

Large language models (LLMs) have demonstrated strong performance across a wide range of natural language processing tasks \citep{Ouyang_TIHF, Sanh_0STG, Wei_0SL, LLAMA, MISTRAL}, including machine translation (MT) \citep{LLM_PROMPT_MT, LLM_HumanLike_Translation, GPT_MT}. 
Recent work \citep{TowerLLM, XU_PARADIGM} has convincingly shown that LLM-based systems consistently outperform specialized encoder-decoder MT systems for many language pairs \citep{WMT24, WMT24++}. 
However, the focus of these studies has been primarily on sentence- and paragraph-level translation, overlooking the complexities of translating entire documents where maintaining coherence, consistency, and discourse structure is fundamental.
For document-level translation, various techniques have emerged, such as context-aware prompting \citep{DocMT_WANG,DocMT_WU} and agent-based translation strategies \citep{AGENT_LLM, DELTA_LLM}. 
On the other hand, supervised fine-tuning (SFT) has proven highly effective for improving sentence-level MT \citep{TowerLLM,XU_PARADIGM}, but its adaptation to document-level translation, and its comparison with other techniques, remain open questions.

\begin{figure*}[!t]
    \centering
    \includegraphics[width=\linewidth]{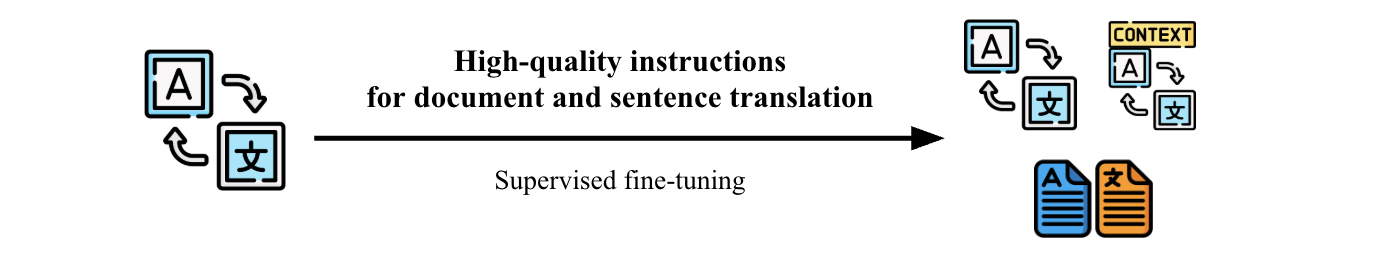}
    \caption{Illustration of our approach for adapting an LLM to document translation.}
    \label{fig:training_example}
\end{figure*}

In this work, we take a fresh perspective on the problem and explore whether we can employ SFT on strong sentence-level LLMs \citep{TowerLLM, XU_PARADIGM, EuroLLM} to adapt them for multilingual contextualization. Specifically, we tackle the underexplored challenge of training LLMs to leverage surrounding context for document-level translation.
Towards this end, we introduce \docblocks, an MT contextualization dataset carefully curated and filtered to incorporate high-quality document-level data. Unlike prior datasets \citep{NLLB, kocmi-etal-2023-findings, WMT24} that focus primarily on isolated sentence pairs or limited context windows, \docblocks consists of full documents and contextual segments sourced from trusted parallel corpora across diverse domains, including news, TED talk transcripts \citep{IWSLT17}, literary texts, and parliamentary proceedings.
We then introduce a simple yet effective contextualization method that fine-tunes existing LLM-based MT systems on \docblocks by integrating the data in multiple formats. These include full documents, contextualized document chunks with preceding sections \citep{DocMT_WANG} and standalone sentence-level examples. This multi-granular training strategy enables the model to capture both document structure and inter-sentence relationships, improving document-level translation performance while maintaining its robust sentence-level capabilities (Figure~\ref{fig:training_example}).
Finally, to evaluate the dependence of our approach on the base MT model, we experiment with three different LLM-based MT models: \textsc{Tower} \citep{TowerLLM}, \textsc{EuroLLM} \citep{EuroLLM}, and \textsc{Qwen2.5} \citep{QWEN_LLM}. This allows us to assess the generalizability of our method across various model architectures.

Our key contributions are summarized as follows:
\begin{itemize}

    \item[\faPuzzlePiece{}] \textbf{\docblocks dataset\footnote{Available on \href{https://huggingface.co/datasets/sardinelab/DocBlocks}{Hugging Face}.}}: We release a document-level parallel dataset for tailoring LLMs to document translation, designed to support long-range dependency modeling and discourse coherence.

    \item[\faCogs{}] \textbf{Fine-tuning for Multilingual Contextualization}: We propose an efficient fine-tuning approach that improves the document-level translation capabilities of LLMs by training on diverse, high-quality instructions across various translation paradigms---improving contextual understanding while preserving sentence-level performance.

    \item[\faSignal{}] \textbf{Comprehensive Evaluation}:
    We conduct a systematic comparison of fine-tuning, prompting, and agent-based methods for document-level MT. Additionally, we perform ablation studies to evaluate the individual components of our approach, demonstrating its effectiveness in tailoring LLMs for document translation.

\end{itemize}

Our experiments across multiple datasets and language pairs demonstrate that our fine-tuning approach, combined with lightweight prompt-based context modeling, significantly improves document-level translation performance across various paradigms, outperforming vanilla LLM-based MT models, as well as both prompting and agent-based techniques.

\section{Multilingual Contextualization of LLMs for Document-level MT}

In this section, we present our contributions, starting with \docblocks, a high-quality parallel dataset curated to address data scarcity in document-level MT (\S\ref{sec:docblocks}). 
Next, we outline the fine-tuning process of LLMs using this dataset to enhance the performance of document-level machine translation (\S\ref{sec:training}).
Finally, we investigate various inference methods and assess their impact on document translation quality (\S\ref{sec:inference}).

\subsection{\faPuzzlePiece{} \docblocks: A High-Quality Document-level MT Parallel Corpora}
\label{sec:docblocks}

\begin{figure}[!ht]
\centering
\footnotesize
\begin{minipage}{\textwidth}
    \centering
    \footnotesize
    \renewcommand{\arraystretch}{1.2}
    \scalebox{0.76}{
        \begin{tabular}{cccccccccccc}
            \toprule
            \multirow{2}{*}{\textbf{Domain}} & \multirow{2}{*}{\textbf{Corpus}} & 
            \multicolumn{4}{c}{\textbf{Pre-filtering}} & \multicolumn{4}{c}{\textbf{Post-filtering}} \\
            \cmidrule(lr){3-6} \cmidrule(lr){7-10}
            & & \textbf{|D|} & \textbf{|S|} & \textbf{|W|} & \textbf{|W|/|D|} & \textbf{|D|} & \textbf{|S|} & \textbf{|W|} & \textbf{|W|/|D|} \\
            \midrule
            \textit{news} & News Commentary v18.1 & 125.0K & 4.9M & 119.0M & 1.0K & 110.0K & 4.4M & 96.4M & 0.9K \\
            \textit{TED Talks} & IWSLT2016/2017 & 33.5K & 3.3M & 82.4M & 2.5K & 29.7K & 2.6M & 64.0M & 2.1K \\
            \textit{Parliamentary} & Europarl v10 & 44.8K & 22.3M & 542.0M & 12.1K & 21.3K & 10.5M & 264.8M & 12.4K \\
            \multirow{2}{*}{\textit{Novels}} & GuoFeng & 22.6K & 1.9M & 51.4M & 2.3K & 18.2K & 1.7M & 28.5M & 1.6K \\
            & BWB & 196.0K & 10.3M & 450.0M & 2.3K & 62.3K & 4.2M & 110.3M & 1.8K \\
            \bottomrule
        \end{tabular}
    }
\end{minipage}
\hfill
\vspace{1em}
\begin{minipage}{\textwidth}
    \centering
    \includegraphics[width=1\linewidth]{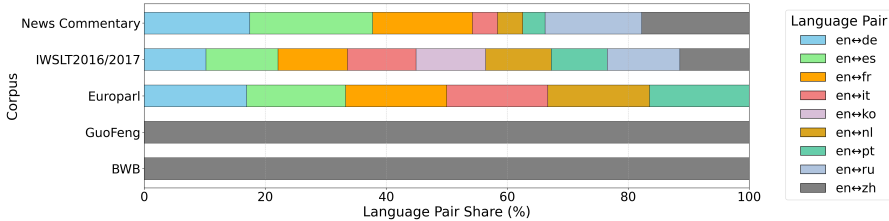}
\end{minipage}
\caption{
    Statistics of datasets used to create \docblocks, including domain, corpus name, number of documents \(|D|\), sentences \(|S|\), words \(|W|\), and average document length \(|W|/|D|\) as a proxy for discourse complexity \citep{DocMT_WANG}. 
    We also illustrate the distribution of language pairs across the filtered datasets, highlighting their multilingual composition.
}
\label{tab:docblocks_aggregated}
\end{figure}

\paragraph{Corpora Collection.}
The \docblocks corpus is constructed from a collection of publicly available document-level datasets, selected to represent a broad range of document types and content domains. 
Figure~\ref{tab:docblocks_aggregated} summarizes the source datasets and their key characteristics. The primary objective of developing \docblocks was to compile a versatile corpus that enables effective training of LLMs for document-level MT tasks.
The \textbf{\textit{News Commentary}} corpus contains political and economic news texts and was included as a representative of well-structured written content from the journalistic domain.
The \textbf{\textit{IWSLT}} \citep{IWSLT17} corpus, based on TED Talk transcripts, provides examples of conversational and spoken language across diverse topics.
The \textbf{\textit{Europarl}} \citep{koehn-2005-europarl} corpus consists of European Parliament proceedings and represents formal writing with longer sentence structures and extended document length, typical of institutional discourse.
The \textbf{\textit{BWB}} \citep{BLONDE} and \textbf{\textit{GuoFeng Webnovel}} \citep{GuoFeng_ST} corpora were included for their expert translations of Chinese novels and web fiction, which feature complex discourse structures and genre diversity.
We also considered the \textbf{\textit{United Nations Parallel Corpus}} \citep{UNPC} and \textbf{\textit{OpenSubtitles}} \citep{OpenSub} for their multilingual scope and size, but document segmentation, alignment issues, and increased computational costs led us to prioritize better-curated datasets.

\paragraph{Data Curation: Preprocessing, Cleaning, and Augmentation.}
All datasets consist of document-level parallel corpora. We do not enforce strict sentence-to-sentence alignment. Instead, alignment is maintained at the document level, allowing for variation in sentence count and structure \---\ reflecting realistic conditions in document translation. This approach enables the model to rely on broader semantic context rather than rigid sentence mappings.
For datasets containing paragraph-level content with document boundary metadata, we grouped paragraphs into full documents and concatenated them up to a maximum length of $32,768$ tokens. To ensure high-quality data, we applied a rigorous cleaning pipeline to \docblocks. Low-quality translations were identified using automatic filters, including Bicleaner \citep{bicleaner} (threshold: $0.5$) and CometKiwi-23 \citep{cometkiwi23} (threshold: $0.65$). We removed entire documents if over $20\%$ of their sentences fell below these thresholds. We further applied language identification with \texttt{langid} \citep{lui-baldwin-2012-langid} to exclude misaligned language pairs. Additional filters removed documents with fewer than $50$ words, more than $10\%$ identical characters/words/numbers, or a source–target length ratio above $1.3$. We also deduplicated all entries.

These steps yielded a clean, reliable dataset optimized for document-level translation. To further enhance learning, we incorporated two complementary training techniques:

\begin{enumerate}
    \item \textit{Multi-Resolutional Document-to-Document Training (MRD2D)} \citep{sun-etal-2022-rethinking}: We split each document into \( k \) parts where \( k \in \{1, 2, 4\} \) to create sequences of varying lengths,  improving computational efficiency.
    \item \textit{Context-Aware Prompt Tuning (CAPT)} \citep{DocMT_WANG}: Building on the same preprocessed chunks, we incorporate a context window of up to the previous $3$ segments in each training prompt. This allows the model to better capture document-level dependencies during training.
\end{enumerate}

We applied MRD2D and CAPT selectively to datasets with longer average document lengths and moderate size, specifically, IWSLT, Europarl, and GuoFeng, to manage data growth and training costs. Experiments show that this curation and training strategy significantly improves translation performance (see Table~\ref{tab:docblocks_ablation}).

\subsection{Document-Level Fine-Tuning with \docblocks}
\label{sec:training}

\paragraph{Model Training.}
Following prior work \citep{ME-1, DocMT_WU}, we adopt a two-step learning strategy, where LLMs are first tuned for sentence-level MT and then adapted for document-level MT tasks.
Given the availability of sentence-level instruction-tuned LLMs~\citep{TowerLLM,EuroLLM}, we focus only on the second step, which further fine-tunes these models on document-level translation data.
To ensure high-quality translations, we calculate the loss exclusively on target tokens, excluding prompt tokens (source and instruction tokens), thereby preventing penalties for prompt format adherence.

\paragraph{\docblocks instruction format.} We support three instruction formats using the \texttt{chatml} \citep{chatml} template.
Following~\citet{DocMT_WU}, we use a contextual block of up to $N=3$ previous chunks and format the texts as shown in Table~\ref{tab:docblocks_chatml}.
In addition to the chunk-level format, we also include instruction examples for document-to-document and sentence-to-sentence translation tasks. These tasks follow a simpler instruction structure, without the contextual block.

\begin{table}[!ht]
\centering
\footnotesize
\begin{tabular}{p{1cm}p{12cm}}
\toprule
\textbf{User}   & {\textcolor{blue}{\texttt{\textbf{\textless|im\_start|>user}}}} \\
                 & \textcolor{gray}{Context:} \\ 
                 &  \textcolor{gray}{{\textbf{\{source\_lang\}}}: {\textbf{\{source$_1$\}}} \; {\textbf{\{target\_lang\}}}: {\textbf{\{target$_1$\}}}} \\
                 &  \textcolor{gray}{{\textbf{\{source\_lang\}}}: {\textbf{\{source$_2$\}}} \; {\textbf{\{target\_lang\}}}: {\textbf{\{target$_2$\}}}} \\
                 &  \textcolor{gray}{{\textbf{\{source\_lang\}}}: {\textbf{\{source$_3$\}}} \; {\textbf{\{target\_lang\}}}: {\textbf{\{target$_3$\}}}} \\
                 & Translate the following source text from {\textbf{\{source\_lang\}}} into {\textbf{\{target\_lang\}}}. \\
                 & {\textbf{\{source\_lang\}}}: {\textbf{\{source\}}}. \\
                 & {\textbf{\{target\_lang\}}}: {\textcolor{blue}{\texttt{\textbf{\textless|im\_end|>}}}} \\
                 & {\textcolor{blue}{\texttt{\textbf{\textless|im\_start|>assistant}}}} \\
\textbf{Model}   & {\textbf{\{target\}}}. {\textcolor{blue}{\texttt{\textbf{\textless|im\_end|>}}}} \\
\bottomrule
\end{tabular}
\caption{Illustration of a \docblocks instance, with the contextual block in \textcolor{gray}{gray}.}
\label{tab:docblocks_chatml}
\end{table}

\subsection{Inference}
\label{sec:inference}
We investigate the impact of different inference methods on document translation performance, as our models are designed to support a variety of translation strategies.
We explore two methods: \textbf{Document-to-Document (Doc2Doc)} and \textbf{Chunking}.

The \textit{Doc2Doc} method translates the entire document in a single pass, leveraging the ability of the LLM to capture long-range context. When the document fits within the model context window, it typically yields the most coherent and consistent translations \citep{DocMT_WANG}.

The \textit{chunking} method translates a document chunk by chunk, with each chunk comprising a fixed number of sentences, paragraphs, or tokens.
It can be applied on its own or enhanced in two complementary ways that address different aspects of the translation process. First, \textit{contextual chunking} improves coherence by conditioning the translation of each chunk on previously translated source-target chunk pairs. Second, \textit{quality-aware chunking}
focuses on selecting the best translation using Minimum Bayes Risk (MBR) decoding \citep{kumar-byrne-2004-minimum,eikema-aziz-2022-sampling}, guided by quality estimation metrics such as \textsc{COMET} \citep{COMET22} and \textsc{ContextCOMET} \citep{vernikos-etal-2022-embarrassingly}. These metrics may incorporate contextual information and help choose the final output by maximizing expected utility:
$
\hat{y}_{mbr} = \arg \max_{y \in \bar{\mathcal{Y}}} \mathbb{E}_{{Y}} \left[ u({Y}, y) \right] 
\approx \arg \max_{y \in \bar{\mathcal{Y}}} \frac{1}{|\bar{\mathcal{Y}}|} \sum_{y_t \in \bar{\mathcal{Y}}} \mathcal{M}([c_t, y_t],[c, y])$, where $\bar{\mathcal{Y}}$ is a list of candidates sampled by the model and $\mathcal{M}$ is the reference-based quality metric, which can optionally use context $c$.

\section{Experiments}
\label{sec:experiments}

\subsection{Experimental Setup}

\paragraph{Datasets.}
We evaluate translation performance using several benchmarks. For document-level MT, we use IWSLT2017 \citep{IWSLT17} and GuoFeng \citep{GuoFeng_ST} test sets. IWSLT2017 contains parallel TED talk documents across six language pairs: en$\leftrightarrow$\{de, fr, it, ko, nl, zh\}. GuoFeng, a discourse-rich web novel corpus, is used for zh$\rightarrow$en experiments on its combined simple and difficult test sets.
For sentence-level MT, we use \textsc{FLORES-200} (en$\leftrightarrow$\{de, fr, it, ko, nl, zh, ru, pt, es\}) \citep{NLLB}, \textsc{WMT23} (en$\leftrightarrow$\{de, ru, zh\}) \citep{kocmi-etal-2023-findings}, and \textsc{TICO-19} (en$\rightarrow$\{es, fr, pt, ru, zh\}) \citep{anastasopoulos-etal-2020-tico} to assess how document-level training affects sentence-level MT quality.

\paragraph{Baselines.}
In this work, we employ the closed-source \textsc{GPT-4o} \citep{GPT-4} and the open-source models \textsc{Qwen2.5-72B-Instruct} \citep{QWEN_LLM} and \textsc{Llama-3.3-70B-Instruct} \citep{grattafiori2024llama3herdmodels} as our baselines.
These three state-of-the-art LLMs serve as the primary comparisons for evaluating our approach to adapting sentence-level LLMs to document-level MT. 
We use greedy decoding for all inference methods, except for quality-aware chunking, where we generate $32$ candidates using $p$-nucleus sampling \citep{2019arXiv190409751H} with \( p = 0.6 \).

\paragraph{Base sentence-level LLMs.}
To assess the effectiveness of our method across different base models, we experiment with three instruction-tuned LLMs: \textsc{TowerInstruct-Mistral-7B} \citep{TowerLLM}, \textsc{EuroLLM-9B-Instruct} \citep{EuroLLM}, and \textsc{Qwen2.5-7B-Instruct} \citep{QWEN_LLM}.
While \textsc{TowerInstruct-Mistral-7B} offers a $32768$ context window, it struggles with document-level MT due to its pretraining and instruction tuning, which are focused on short sequences. 
\textsc{EuroLLM-9B-Instruct} has a $4096$ context window, limiting its performance in document-level MT. In contrast, \textsc{Qwen2.5-7B-Instruct} performs better with its $32768$ context window (see Tables~\ref{tab:GuoFeng_doc2doc} and~\ref{tab:iwslt_doc2doc}).

\paragraph{Hyperparameters.} We fine-tune all sentence-level LLMs on the instructional \docblocks using the standard cross-entropy loss and enable \texttt{bfloat16} mixed precision and packing \citep{raffel2020exploring}.  
We detail all hyperparameters in Appendix~\ref{sec:docmt_hyperparams}.
For \textsc{TowerInstruct-Mistral-7B} and \textsc{Qwen2.5-7B-Instruct}, we kept the original RoPE theta \citep{ROPE} since both were pretrained with a $32768$ context window. For EuroLLM, initial tests showed minimal gains from extending RoPE, so we retained the original design.

\paragraph{Evaluation.}
We evaluate different approaches using both sentence-level and document-level metrics. Sentence-level evaluation includes \textsc{BLEU} \citep{BLEU} and \textsc{COMET} (\texttt{wmt22-comet-da}) \citep{COMET22}.
Sentence alignments are first computed using \texttt{bleualign}, after which these metrics are applied. 
However, these metrics have known limitations in capturing document-level phenomena such as coherence, cohesion, and referential consistency \citep{CASTILHO_SPAN,FERNANDES_MEASURING,LAUBLI,TORAL,FREITAG_EEC}.
To better evaluate document-level translation quality, we report document-level \textsc{BLEU} (d-\textsc{BLEU}) \citep{BLEU,liu-etal-2020-multilingual-denoising} with corresponding brevity penalty (\textsc{BP}), and document-level \textsc{COMET} (d-\textsc{COMET}) \citep{COMET22}, computed using the SLIDE approach \citep{raunak-etal-2024-slide}. SLIDE divides each document into overlapping 512-token chunks (the context limit for \textsc{COMET}), computes chunk-level scores independently, and averages them to obtain a document-level score.
We also report the Lexical Translation Consistency Ratio (\textsc{LTCR}) \citep{DocMT_WANG,DELTA_LLM,xiao-etal-2011-document,lyu-etal-2021-encouraging}, which captures the consistency of repeated terminology across a document. Although recent work has proposed alternative document-level metrics \citep{vernikos-etal-2022-embarrassingly,BLONDE,DocMetric-1,DocMetric-2}, there remains no universally accepted method for evaluating document-level translation quality. 
Developing more robust and comprehensive evaluation strategies remains an important open problem, complementary to ours, which we leave for future work.

\subsection{Results and Analysis}

\paragraph{DocMT-LLMs achieve superior document-to-document (Doc2Doc) translation quality.}
DocMT-LLMs (LLM-based document-level MT models) consistently outperform sentence-level models across all benchmarks. 
On both GuoFeng (Table~\ref{tab:GuoFeng_doc2doc}) and IWSLT2017 (Table~\ref{tab:iwslt_doc2doc}) datasets, DocMT-LLMs all outperform their sentence-level counterparts, with strong gains across both sentence- and document-level metrics.
These lower scores for sentence-level baselines are expected, as they lack access to document-wide context during training, which is crucial for ensuring translation coherence.
Despite being much smaller, our models outperform strong baselines like \textsc{Llama-3.3-70B-Instruct}, \textsc{Qwen2.5-72B-Instruct} and \textsc{GPT-4o} on GuoFeng and perform well on IWSLT2017. While these baselines slightly lead on d-\textsc{COMET} and on d-\textsc{BLEU} (only for IWSLT2017 en$\rightarrow$xx), our models still achieve substantial gains over undertrained sentence-level LLMs.
These results demonstrate that incorporating \docblocks enables sentence-level LLMs to generate coherent, high-quality document-level translations.

\begin{table}[!htbp]
\centering
\renewcommand{\arraystretch}{1.1}
\resizebox{0.8\textwidth}{!}{
\begin{tabular}{lcccccc}
\toprule
\multirow{2}{*}{\textbf{Models}} & \multicolumn{5}{c}{\textbf{GuoFeng zh$\rightarrow$en}} \\\cmidrule{2-6}
&\textbf{\textsc{BLEU}} &\textbf{\textsc{COMET}} &\textbf{d-\textbf{BLEU} (BP)} &\textbf{d-\textsc{COMET}} &\textbf{\textsc{LTCR}} \\\midrule
\texttt{Llama-3.3-70B-Instruct} & 12.12 & 80.16 & 15.32 (0.83) & 71.37 & 65.24 \\
\texttt{Qwen2.5-72B-Instruct} & \underline{17.87} & \textbf{\underline{84.85}} & \underline{20.99 (0.97)} & 72.13 & \underline{65.72} \\
\texttt{GPT-4o} &14.61 &73.66 &17.46 (1.00) & \textbf{\underline{73.73}} & 59.69 \\
\midrule
\texttt{TowerInstruct-Mistral-7B} &6.78 &58.40 &7.27 (1.00) &58.23 &54.16 \\
\texttt{DocMT-TowerInstruct-Mistral-7B} & \textbf{\underline{34.06}} & \underline{78.72} & \textbf{\underline{37.57 (1.00)}} & \underline{72.84} & \textbf{\underline{74.59}} \\
\midrule
\texttt{EuroLLM-9B-Instruct} &12.43 &63.14 &12.03 (0.84) &60.11 &51.07 \\
\texttt{DocMT-EuroLLM-9B-Instruct} & \underline{24.12} & \underline{73.88} & \underline{27.83 (1.00)} & \underline{72.41} & \underline{67.05} \\
\midrule
\texttt{Qwen2.5-7B-Instruct} & 16.51 & \underline{84.34} & 19.56 (0.96) & 71.66 & 63.22 \\
\texttt{DocMT-Qwen2.5-7B-Instruct} & \underline{30.67} & 77.28 & \underline{34.80 (0.99)} & \underline{71.88} & \underline{73.48} \\
\bottomrule
\end{tabular}
}
\caption{Doc2Doc translation results on the GuoFeng dataset. Best overall values are \textbf{bolded}, and group-specific bests are \underline{underlined}.
}
\label{tab:GuoFeng_doc2doc}
\end{table}

\vspace{-0.07cm}
\begin{table}[!htbp]
\centering
\setlength{\tabcolsep}{2pt}
\resizebox{1\textwidth}{!}{
\begin{tabular}{l *{5}{c} | *{5}{c}}
\toprule
\multirow{2}{*}{\textbf{Models}} & \multicolumn{5}{c}{\textbf{IWSLT2017 en$\rightarrow$xx}} &\multicolumn{5}{c}{\textbf{IWSLT2017 xx$\rightarrow$en}} \\\cmidrule{2-6} \cmidrule{7-11}
&\textbf{\textsc{BLEU}} &\textbf{\textsc{COMET}} &\textbf{d-\textbf{BLEU (BP)}} &\textbf{d-\textsc{COMET}} &\textbf{\textsc{LTCR}} &\textbf{\textsc{BLEU}} &\textbf{\textsc{COMET}} &\textbf{d-\textbf{BLEU (BP)}} &\textbf{d-\textsc{COMET}} &\textbf{\textsc{LTCR}} \\\midrule
\texttt{Llama-3.3-70B-Instruct} & 15.61 & 61.17 & 27.08 (0.85) & 70.56 & 57.64 & 12.34 & 54.87 & 37.91 (0.90) & 72.01 & \underline{82.20} \\
\texttt{Qwen2.5-72B-Instruct} & 18.48 & 68.47 & 26.59 (0.93) & \textbf{\underline{75.15}} & \underline{60.75} &13.12 &58.85 & \underline{38.81 (0.93)} & 68.44 & 81.57 \\
\texttt{GPT-4o} &\textbf{\underline{27.96}} & \underline{74.79} & \textbf{\underline{32.25 (0.97)}} & 74.73 & 59.05 & \underline{28.53} & \underline{76.55} & 33.47 (0.89) & \textbf{\underline{75.04}} &77.88 \\\midrule
\texttt{TowerInstruct-Mistral-7B} &1.77 &41.67 &2.41 (0.62) &29.68 &27.53 &10.10 &59.64 &2.36 (0.21) &31.79 &28.88 \\
\texttt{DocMT-TowerInstruct-Mistral-7B} &\underline{22.67} & \underline{77.33} &\underline{25.60 (0.94)} &\underline{74.38} &\underline{60.46} &\underline{27.73} &\underline{59.69} &\underline{26.79 (1.00)} &\underline{60.29} &\underline{79.97} \\\midrule
\texttt{EuroLLM-9B-Instruct} &6.51 &73.64 &3.75 (0.56) &24.97 &35.20 &8.83 & \underline{78.94} &3.97 (0.39) &29.81 &33.79 \\
\texttt{DocMT-EuroLLM-9B-Instruct} &\underline{20.94} & \underline{72.73} &\underline{20.24 (1.00)} &\underline{61.74} &\underline{55.47} & \underline{28.75} & 69.66 &\underline{28.46 (1.00)} &\underline{62.71} &\underline{78.20} \\
\midrule
\texttt{Qwen2.5-7B-Instruct} & 17.32 & 70.57 & 20.86 (0.80) & 71.73 & 56.36 & 23.94 & 75.95 & 34.59 (0.90) & \underline{73.29} & 80.50 \\
\texttt{DocMT-Qwen2.5-7B-Instruct} & \underline{23.98} & \textbf{\underline{78.48}} & \underline{25.29 (0.98)} & \underline{73.00} & \textbf{\underline{62.05}} & \textbf{\underline{37.64}} & \textbf{\underline{81.91}} & \textbf{\underline{40.56 (0.97)}} & 73.02 & \textbf{\underline{84.09}}
\\\bottomrule
\end{tabular}
}
\caption{Doc2Doc translation results on IWSLT2017, averaged over en$\rightarrow$xx and xx$\rightarrow$en. Best overall values are \textbf{bolded}, and group-specific bests are \underline{underlined}.
}
\label{tab:iwslt_doc2doc}
\end{table}

\begin{figure}[!htbp]
    \centering
    \begin{minipage}{0.325\textwidth}
        \centering
        \includegraphics[width=\textwidth]{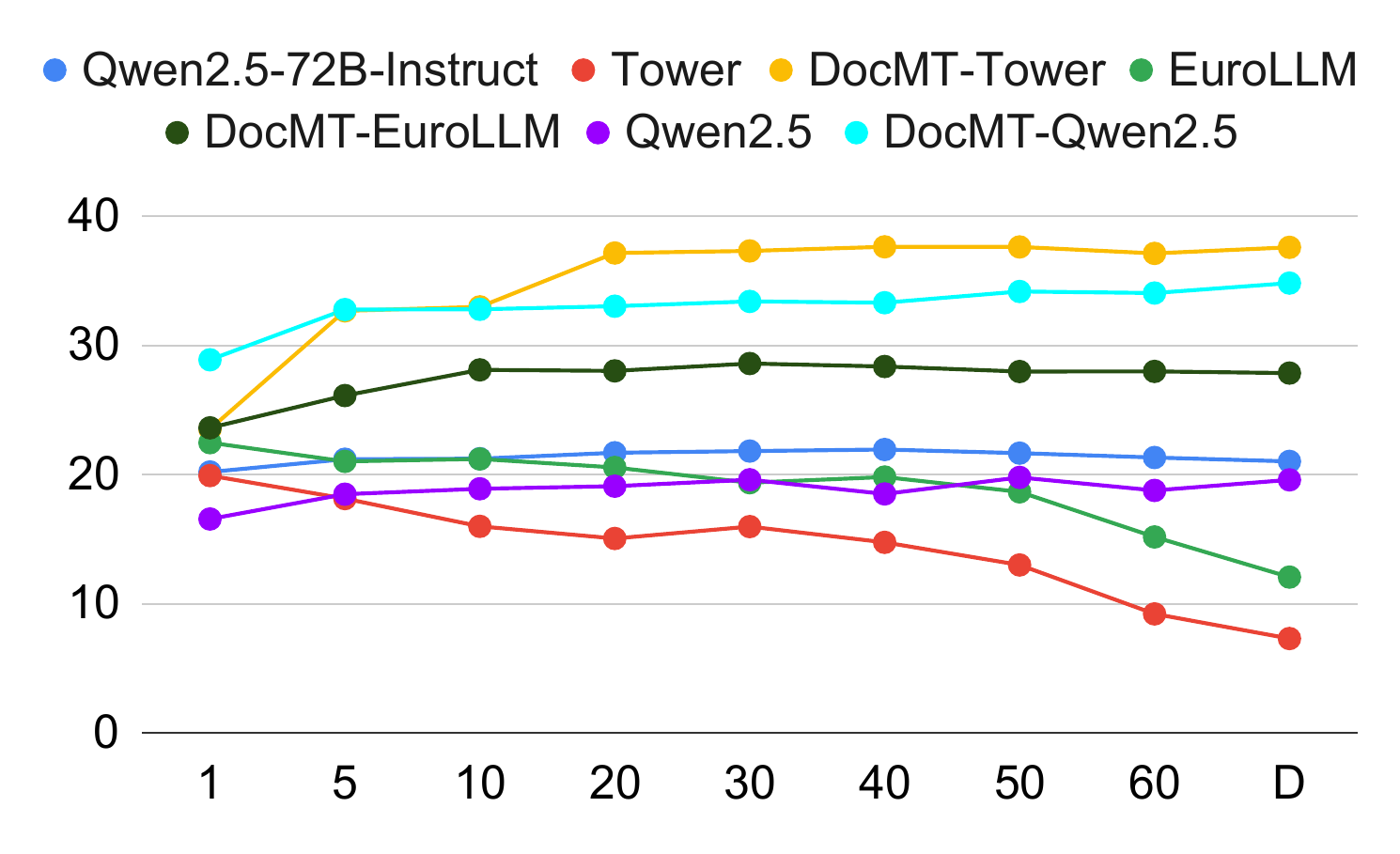}
    \end{minipage} \hfill
    \begin{minipage}{0.325\textwidth}
        \centering
        \includegraphics[width=\textwidth]{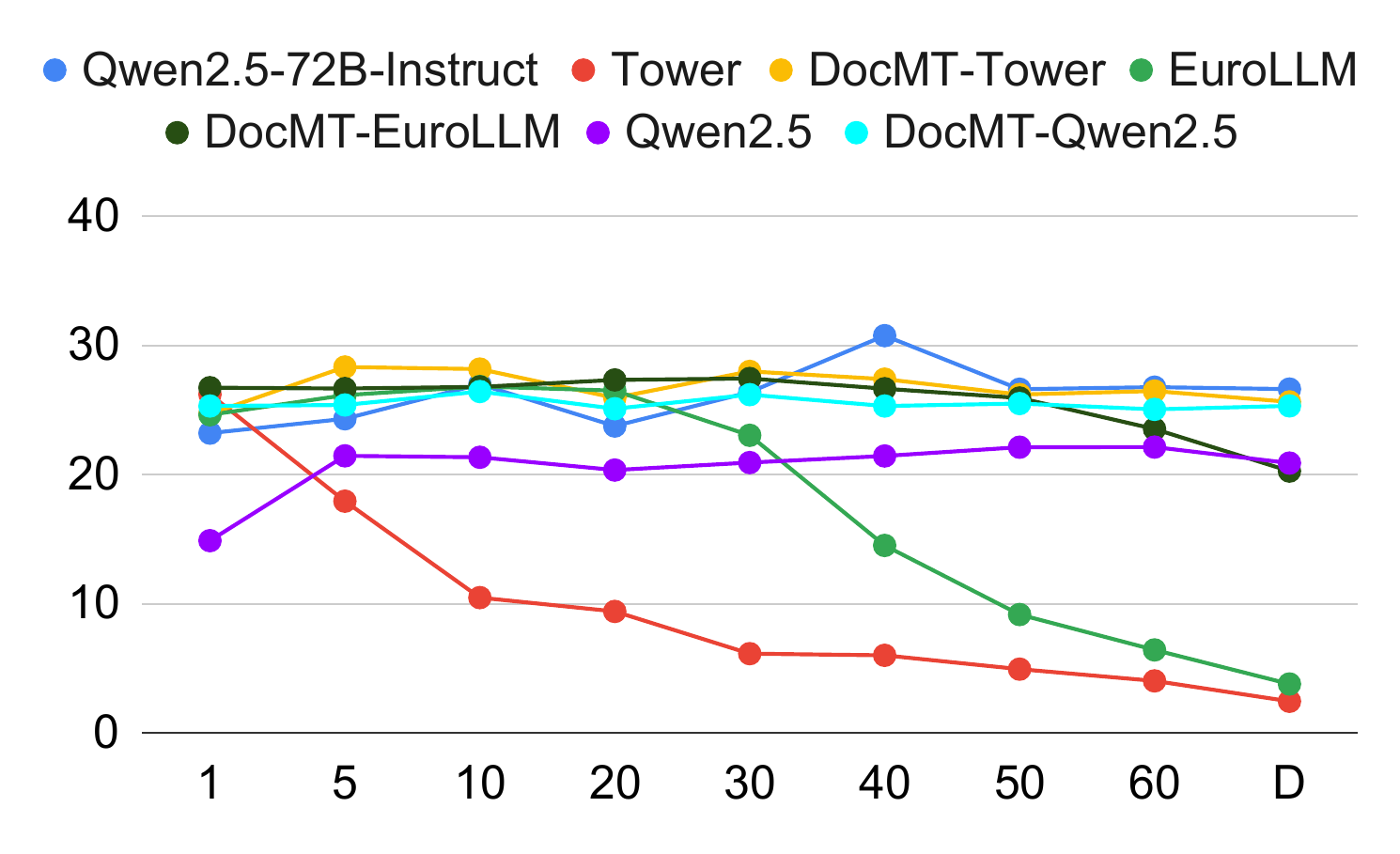}
    \end{minipage} \hfill
    \begin{minipage}{0.325\textwidth}
        \centering
        \includegraphics[width=\textwidth]{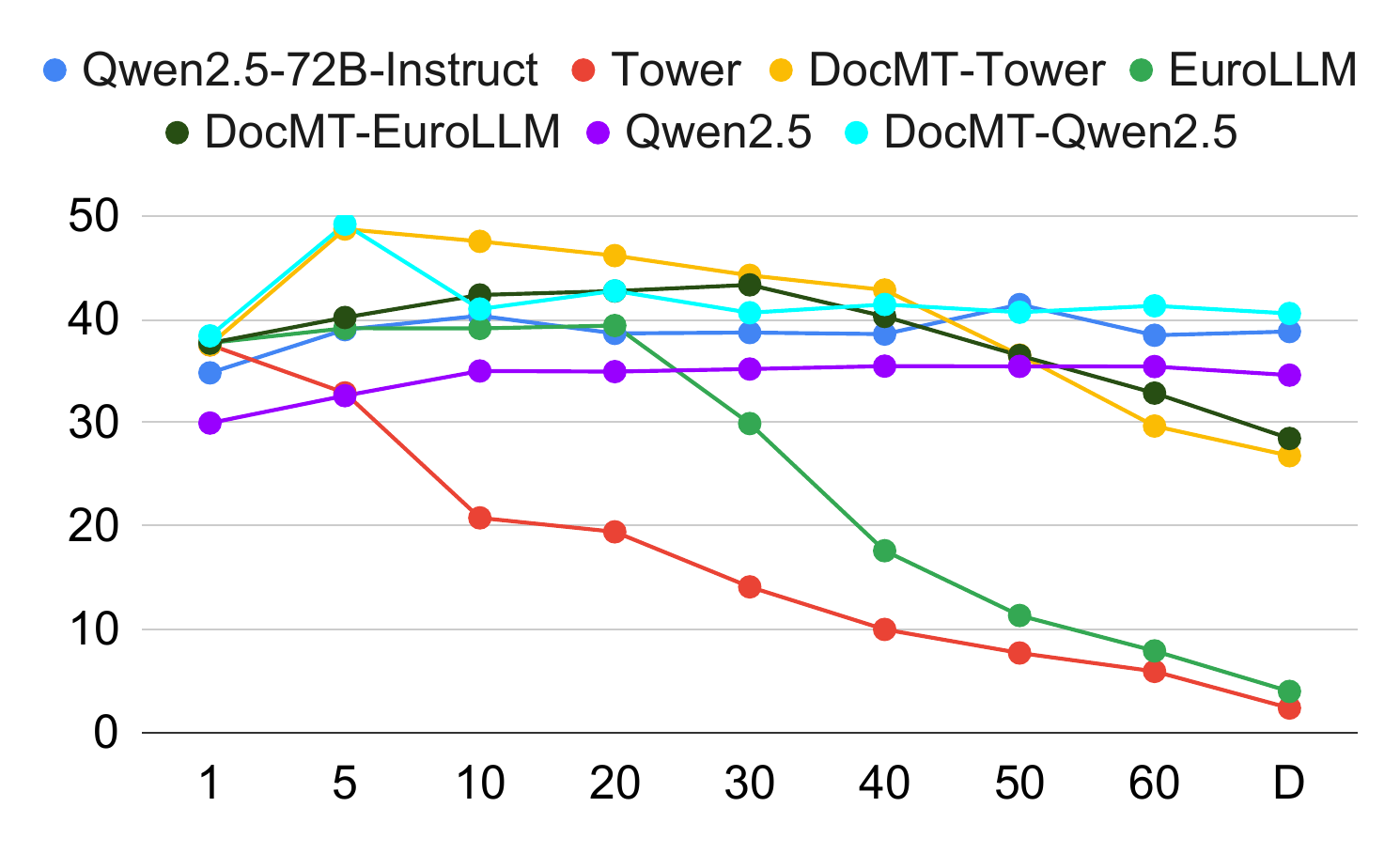}
    \end{minipage}
    \begin{minipage}{0.325\textwidth}
        \centering
        \includegraphics[width=\textwidth]{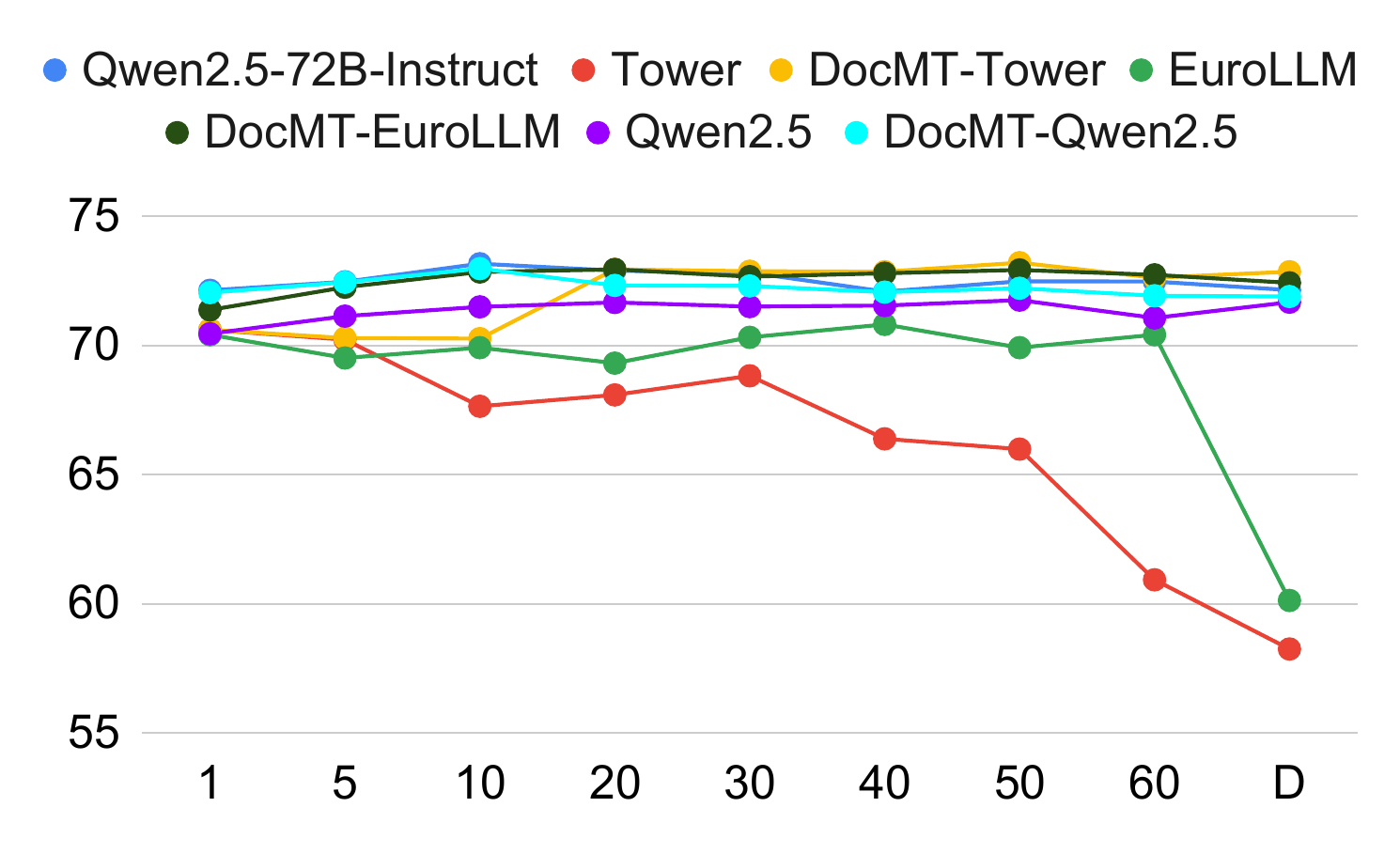}
    \end{minipage} \hfill
    \begin{minipage}{0.325\textwidth}
        \centering
        \includegraphics[width=\textwidth]{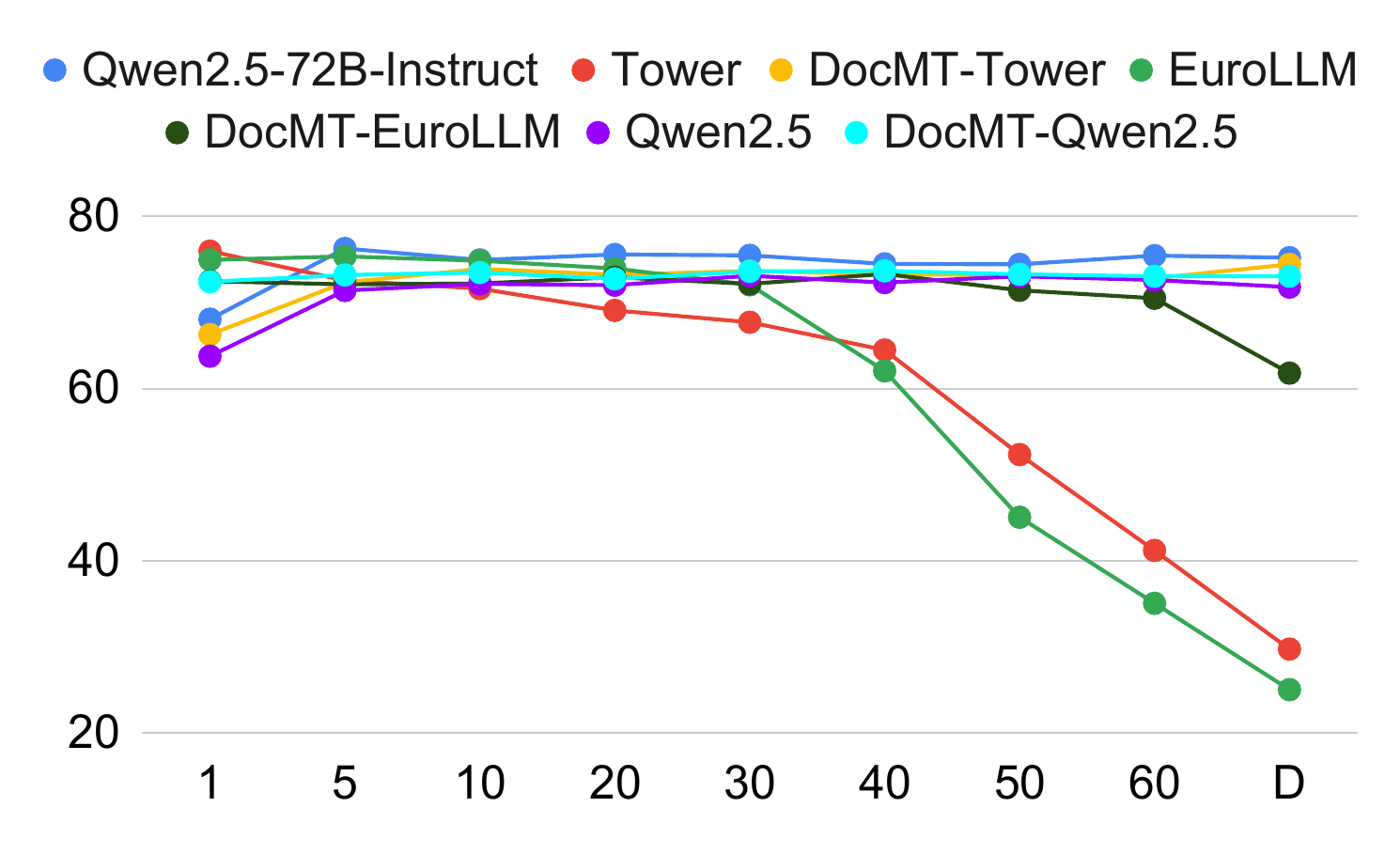}
    \end{minipage} \hfill
    \begin{minipage}{0.325\textwidth}
        \centering
        \includegraphics[width=\textwidth]{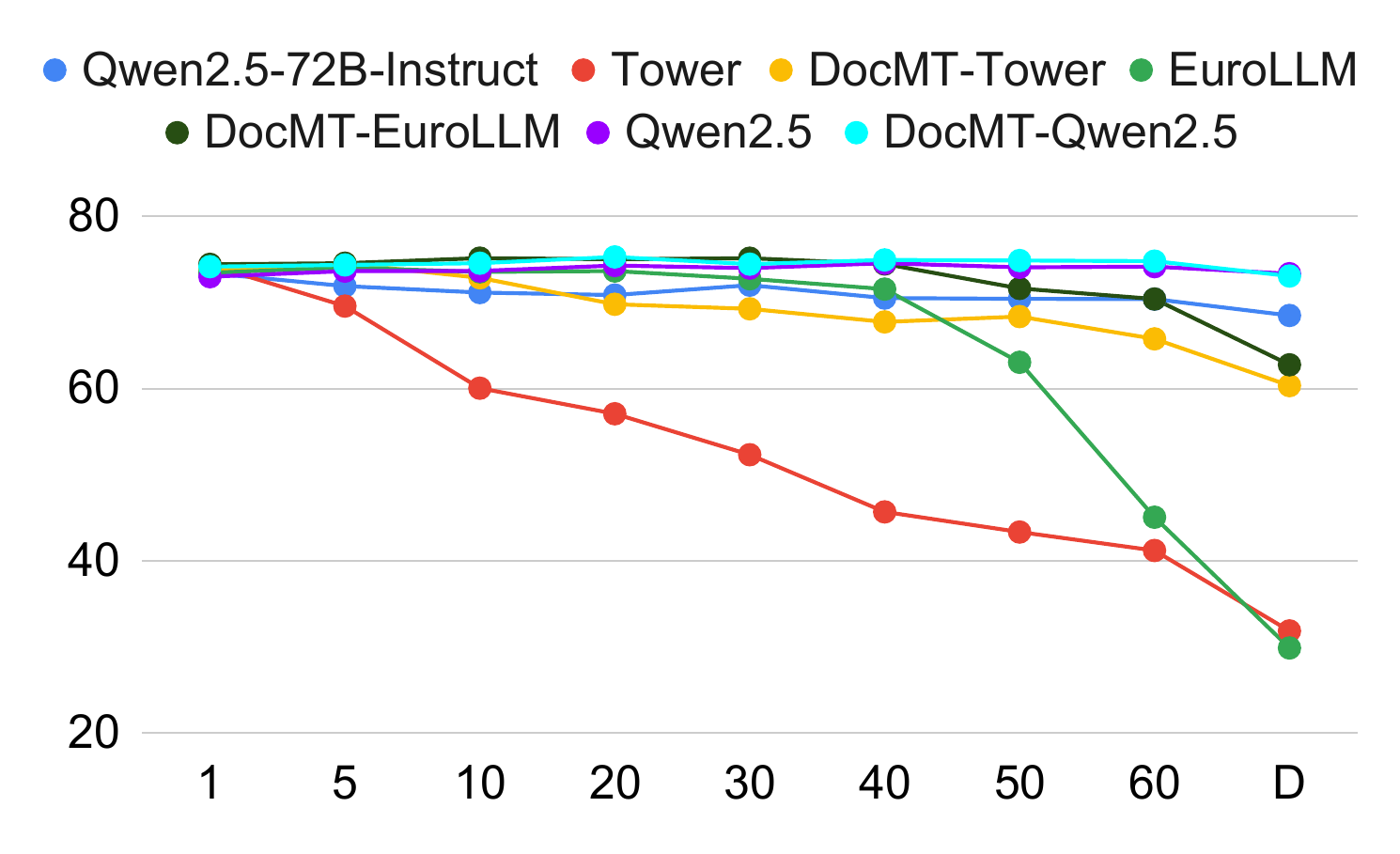}
    \end{minipage}
    \vspace{-0.5em}
    \caption{d-\textsc{BLEU} (top) and d-\textsc{COMET} (bottom) scores for the GuoFeng (left), IWSLT2017 en$\rightarrow$xx (middle), and xx$\rightarrow$en (right) datasets as the decoding chunk size increases. “D” indicates that the entire document is treated as a single chunk during decoding.
    }
    \label{fig:chunking}
\end{figure}

\paragraph{DocMT-LLMs deliver high-quality chunking decoding.}
Chunking-based decoding enables efficient document translation by processing text in segments. Larger chunks provide more context, bridging sentence- and document-level translation. DocMT-LLMs consistently outperform sentence-level models on d-\textsc{BLEU} and d-\textsc{COMET} (Figure~\ref{fig:chunking}), especially at larger chunk sizes. In contrast, \textsc{TowerInstruct-Mistral-7B} and \textsc{EuroLLM-9B-Instruct} degrade as the chunk size increases. Document-level models maintain strong performance and often surpass \textsc{Qwen2.5-72B-Instruct}. \textsc{DocMT-Qwen2.5-7B-Instruct} improves quality across most chunk sizes, despite \textsc{Qwen2.5-7B-Instruct} already supporting document-level MT. These results show that training on \docblocks enhances both document translation and chunking decoding, making DocMT-LLMs more robust.

\begin{figure}[!htbp]
    \centering
    \begin{minipage}{0.5\textwidth}
        \includegraphics[width=\textwidth]{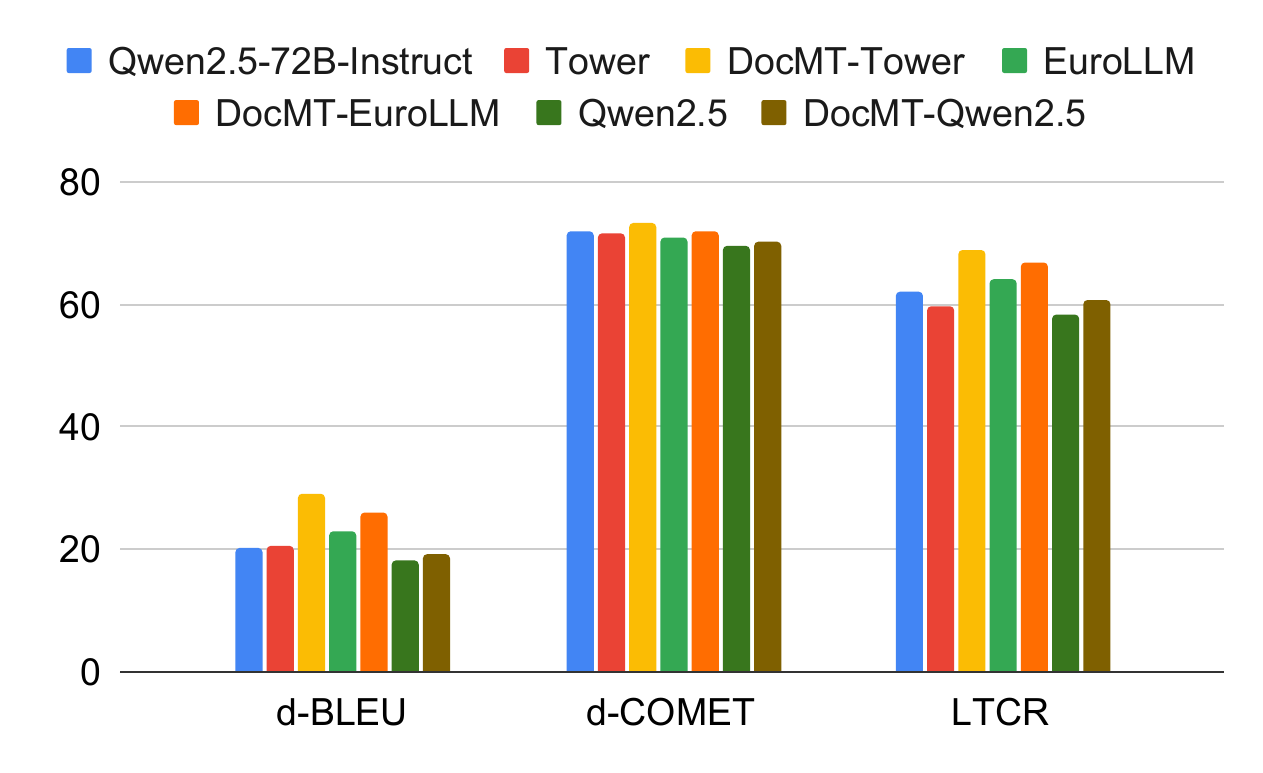}
        \vspace{-1.5em}
        \caption{Evaluation scores for contextual chunking on GuoFeng using a context window of $3$, consistent with the training setup.
        Improvements demonstrate CAPT’s effect in enabling DocMT-LLMs to better leverage prior translated chunks during decoding.
        }
        \label{fig:cap}
    \end{minipage}
    \hspace{\fill}
    \begin{minipage}{0.47\textwidth}
        \includegraphics[width=\textwidth]{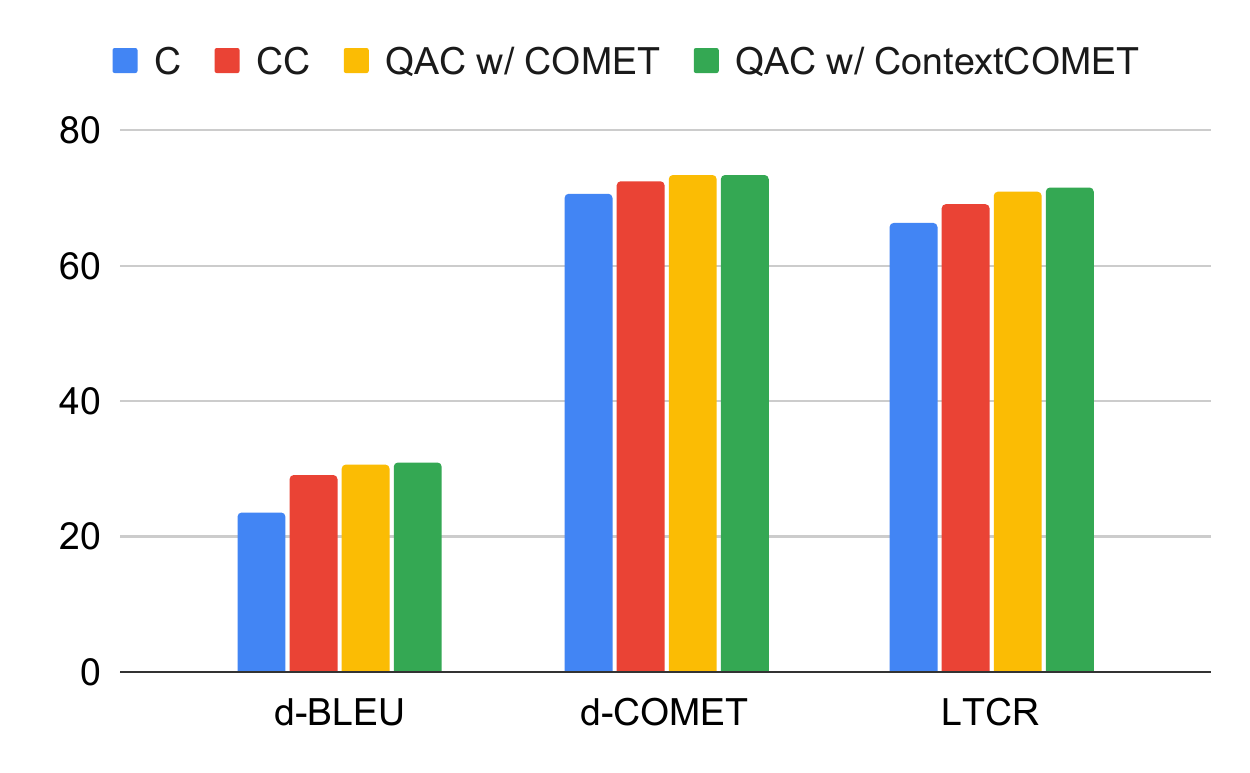}
        \vspace{-1.0em}
        \caption{\textsc{DocMT-TowerInstruct-7B} performance on GuoFeng using chunking (C), contextual chunking (CC), and quality-aware chunking (QAC), which combines contextual chunking with paragraph-level quality metrics.}
        \label{fig:qac}
    \end{minipage}
\end{figure}

\paragraph{DocMT-LLMs can be coupled with contextual chunking and quality-aware chunking to improve translation quality.}
DocMT-LLMs improve translation quality by incorporating context during chunk-level decoding. 
Figure~\ref{fig:cap} shows that context-aware prompt tuning (CAPT) helps models better use surrounding context, leading to higher scores across document-level metrics.
Quality-aware chunking (Figure~\ref{fig:qac}) further improves performance by combining contextual chunking with paragraph-level metrics like \textsc{COMET} and \textsc{ContextCOMET}, using a context window of three chunks within a $512$-token limit. However, such metrics may not fully capture document-level quality and can lead to suboptimal optimization.
As shown in Table~\ref{tab:inference_comparison}, standard chunking is the most efficient due to fully parallelizable decoding. In contrast, contextual and quality-aware chunking require sequential decoding, with quality-aware chunking being slower due to candidate reranking with quality metrics. Nevertheless, \textsc{DocMT-TowerInstruct-Mistral-7B} supports multiple translation paradigms, offering varying translation quality and efficiency trade-offs.

\begin{table}[!htbp]
\centering
\resizebox{\textwidth}{!}{
\begin{tabular}{lcccc}
\toprule
\textbf{Inference Method} & \textbf{Chunking} & \textbf{Contextual Chunking} & \textbf{Quality-aware Chunking} & \textbf{Document-to-Document} \\ \midrule
\textbf{Throughput $\uparrow$} & 392.45 & 39.53 & 16.67 & 204.22 \\ 
\bottomrule
\end{tabular}
}
\caption{
Throughput comparison of inference methods on Guofeng with \textsc{DocMT-TowerInstruct-Mistral-7B}, measured in tokens per second.}
\label{tab:inference_comparison}
\end{table}

\paragraph{Sentence-level MT capabilities of DocMT-LLMs do not deteriorate despite document-level MT training on \docblocks.}
Table~\ref{tab:SentMT} shows that DocMT-LLMs fine-tuned with \docblocks retain strong sentence-level performance, with COMET score differences from their sentence-level counterparts typically within $0.5$ points, indicating no catastrophic forgetting. \textsc{Qwen2.5-7B-Instruct} even improves sentence-level performance through document-level training, enhancing robustness without sacrificing sentence-level quality. 
Detailed results per language pair are provided in Appendix~\ref{sec:full_mt_results}.

\begin{table}[!htbp]
\centering
\footnotesize
\renewcommand{\arraystretch}{1.1}
\resizebox{0.9\textwidth}{!}{
\begin{tabular}{lccccc}
\toprule
\multirow{2}{*}{\textbf{Models}} & \multicolumn{2}{c}{\textbf{FLORES-200}} & \multicolumn{2}{c}{\textbf{WMT23}} & \textbf{TICO-19} \\
                & \textbf{en$\rightarrow$xx} & \textbf{xx$\rightarrow$en} & \textbf{en$\rightarrow$xx} & \textbf{xx$\rightarrow$en} & \textbf{en$\rightarrow$xx} \\
\midrule
\texttt{TowerInstruct-Mistral-7B} & 88.96 & 88.45 & 85.26 & 83.03 & 87.46 \\
\texttt{DocMT-TowerInstruct-Mistral-7B} & 88.54 & 88.21 & 84.93 & 82.97 & 87.17 \\
\midrule
\texttt{EuroLLM-9B-Instruct} & 89.21 & 88.31 & 85.54 & 83.24 & 87.53 \\
\texttt{DocMT-EuroLLM-9B-Instruct} & 88.79 & 88.41 & 85.26 & 82.88 & 87.20 \\
\midrule
\texttt{Qwen2.5-7B-Instruct} & 83.08 & 86.70 & 79.89 & 79.81 & 83.36 \\
\texttt{DocMT-Qwen2.5-7B-Instruct} & 87.77 & 88.06 & 83.91 & 81.80 & 86.51 \\
\bottomrule
\end{tabular}
}
\caption{Evaluation of sentence-level LLMs and DocMT-LLMs based on \textsc{COMET} across sentence-level MT benchmarks.}
\label{tab:SentMT}
\end{table}

\begin{figure}[!htbp]
    \centering
    \includegraphics[width=.99\textwidth]{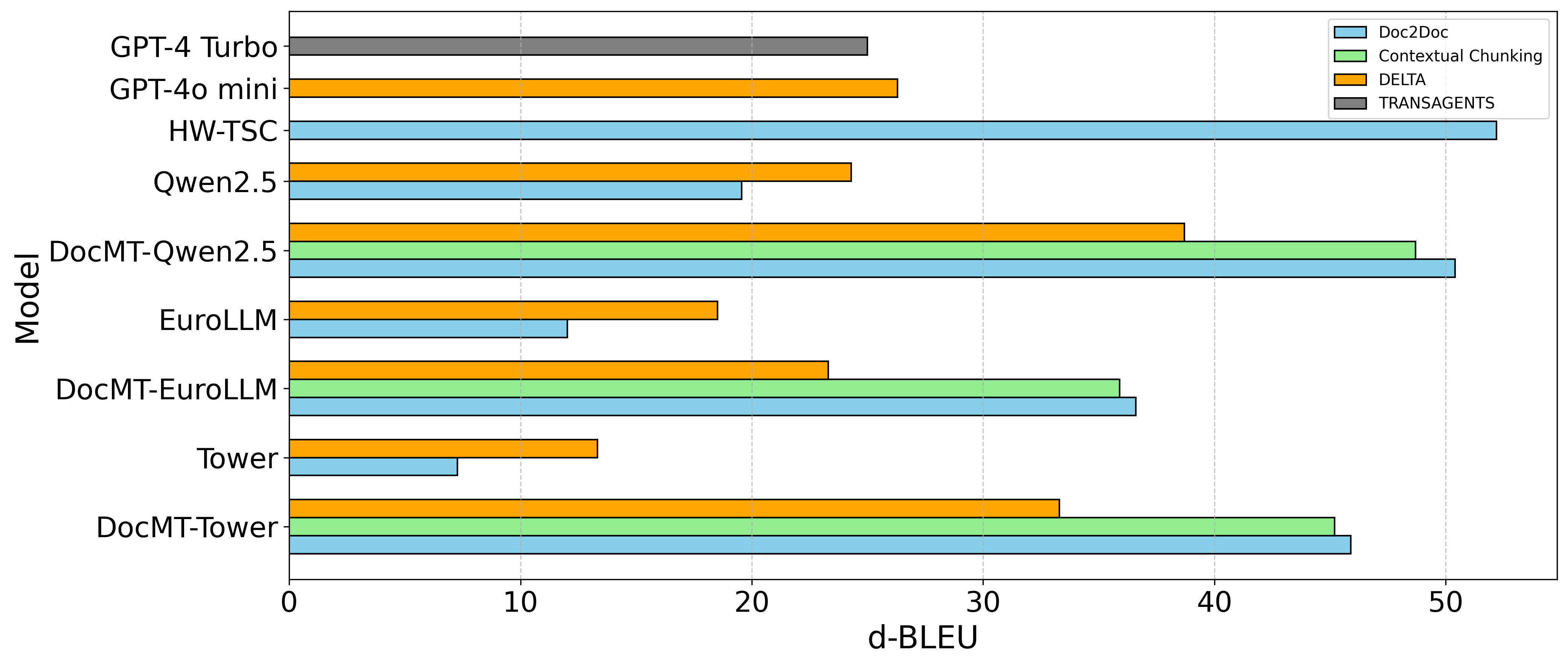}
    \caption{  
    d-\textsc{BLEU} scores on the Guofeng test set, with chapter translations concatenated into a single document \citep{AGENT_LLM}. The figure compares document-level translation quality across different paradigms: blue for Doc2Doc, green for Contextual Chunking, orange for the DELTA agent-based approach, and gray for the TRANSAGENTS agent-based approach. In addition to evaluating our base models before and after DocMT training, we include strong external baselines representative of each paradigm \---\ TRANSAGENTS with GPT-4 Turbo \citep{AGENT_LLM}, DELTA with GPT-4o mini \citep{DELTA_LLM}, and HW-TSC \citep{xie-etal-2023-hw} for Doc2Doc.
    }
    \label{fig:DocMTvsDELTA}
\end{figure}

\paragraph{Comparing translation approaches: DocMT-LLMs vs. Agent-based methods.}
DocMT-LLMs consistently outperform (in quality and inference speed) agent-based methods like TRANSAGENTS \citep{AGENT_LLM} and DELTA \citep{DELTA_LLM} in both Doc2Doc and contextual chunking translation tasks  (Figure~\ref{fig:DocMTvsDELTA}).
Agent-based methods iteratively refine translations using memory-based mechanisms with dictionaries and using a hierarchical workflow. This multi-step pipeline and inter-agent communication can introduce significant latency, slowing inference.
In contrast, training a DocMT-LLM requires $120$ hours on $4$ NVIDIA H$100$ $80$GB GPUs but runs efficiently on a single NVIDIA A$6000$ $48$GB GPU afterward. This initial cost is offset after processing just 42 documents of 400 sentences each.
Therefore, both contextual chunking and Doc2Doc approaches with DocMT-LLMs achieve a superior balance among context preservation, speed, scalability, and translation quality.

\paragraph{\docblocks curation process impacts learning.}
As shown in Table~\ref{tab:docblocks_ablation}, filtering improves model performance across most datasets. The Multiresolution Document-to-Document (MRD2D) technique provides significant benefits for both document-to-document translation and chunking by introducing additional subdocuments during training, especially for the IWSLT2017 dataset. Context-Aware Prompt Tuning (CAPT) is particularly effective for Contextual Chunking, as it provides context-driven examples that improve performance, particularly for the GuoFeng and IWSLT2017 (xx$\rightarrow$en) tasks.
Combining filtering with MRD2D and CAPT yields the best results overall, boosting translation quality and contextual understanding across document-level translation tasks.
We further examine the impact of balancing the \docblocks dataset, which is naturally skewed toward English-to-Chinese document pairs due to their greater availability. In this setting, we uniformly sample both from and into English across language directions to ensure an even distribution of translation pairs. Our results indicate that enforcing this balance yields no performance gains. As such, we choose to retain as much data as possible rather than discard samples to achieve balance, prioritizing overall model performance.

\begin{table}[!htbp]
\renewcommand{\arraystretch}{1.3}
\centering
\setlength{\tabcolsep}{2pt}
\resizebox{\textwidth}{!}{
\begin{tabular}{lccccccccc}
\toprule
\multirow{3}{*}{\textbf{Models}} & \multicolumn{3}{c}{\textbf{Doc2Doc}} & \multicolumn{3}{c}{\textbf{Chunking}} & \multicolumn{3}{c}{\textbf{Contextual Chunking}} \\
\cmidrule(lr){2-4} \cmidrule(lr){5-7} \cmidrule(lr){8-10}
 & \textbf{GuoFeng} & \multicolumn{2}{c}{\textbf{IWSLT2017}} & \textbf{GuoFeng} & \multicolumn{2}{c}{\textbf{IWSLT2017}} & \textbf{GuoFeng} & \multicolumn{2}{c}{\textbf{IWSLT2017}} \\
 & zh$\rightarrow$en & en$\rightarrow$xx & xx$\rightarrow$en & zh$\rightarrow$en & en$\rightarrow$xx & xx$\rightarrow$en & zh$\rightarrow$en & en$\rightarrow$xx & xx$\rightarrow$en \\
\midrule
\textsc{TowerInstruct-Mistral-7B} & 58.23 & 29.68 & 31.79 & 68.81 & 67.61 & 52.26 & 70.21 & 70.62 & 69.53 \\
\midrule
\multicolumn{10}{l}{\textbf{SFT}}\vspace{0.5mm}  \\
\footnotesize{+ unfiltered \docblocks} & 70.08 & 62.30 & 55.32 & 69.20 & 69.24 & 55.89 & 70.55 & 69.34 & 69.00 \\
\footnotesize{+ filtered \docblocks} & 71.71 & 63.67 & 57.72 & 69.21 & 70.36 & 58.23 & 70.45 & 69.77 & 68.44 \\
\footnotesize{~\quad + MRD2D} & 72.55 & \textbf{74.60} & 60.11 & 72.38 & 73.03 & 68.97 & 70.34 & 71.32 & 69.45 \\
\footnotesize{~\quad + CAPT} & 71.70 & 66.33 & 58.31 & 68.80 & 68.55 & 65.37 & \textbf{73.11} & 73.05 & \textbf{75.04} \\
\footnotesize{~\quad + MRD2D + CAPT} & \textbf{72.84} & 74.38 & \textbf{60.29} & \textbf{72.86} & \textbf{73.51} & \textbf{69.21} & 72.93 & \textbf{73.18} & 74.48 \\
\footnotesize{+ balanced \docblocks} & 67.89 & 61.79 & 55.03 & 65.72 & 69.57 & 54.39 & 69.80 & 70.33 & 68.90 \\

\bottomrule
\end{tabular}
}
\caption{Ablation results for the components of \docblocks, reporting d-\textsc{COMET} scores. 
}
\label{tab:docblocks_ablation}  
\end{table}

\begin{figure}[!htbp]
    \centering
    \begin{minipage}{0.496\textwidth}
        \includegraphics[width=\textwidth]{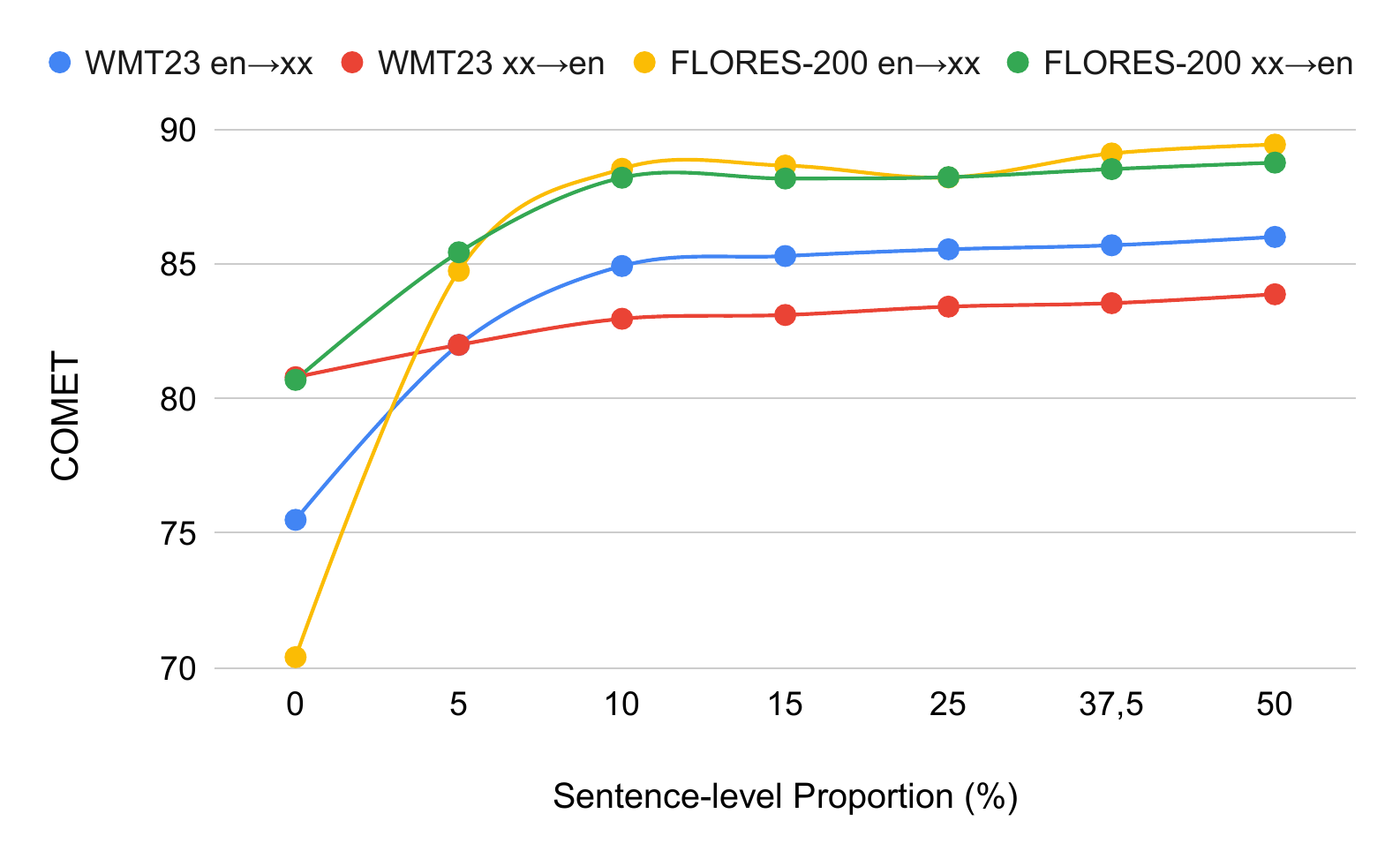}
    \end{minipage} \hfill
    \begin{minipage}{0.496\textwidth}
        \includegraphics[width=\textwidth]{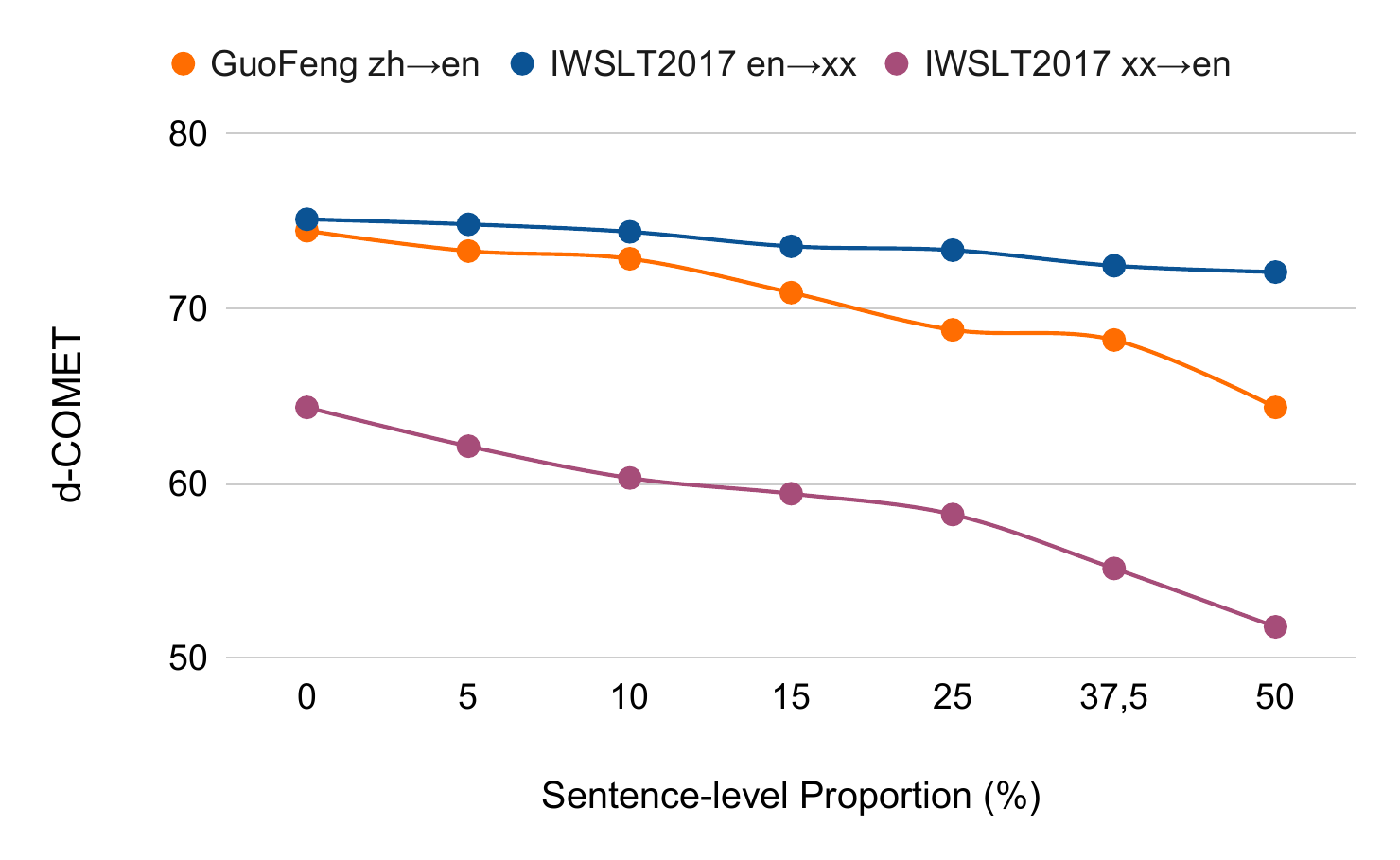}
    \end{minipage}
    \caption{Effect of varying sentence-level data in \docblocks training on sentence-level (left) and document-level (right) translation performance.
    }
    \label{fig:sent_doc_prop}
\end{figure}

\paragraph{Balancing sentence- and document-Level data in \docblocks: impacts on translation performance.}
We conducted an ablation study to assess how varying the proportion of sentence-level data in \docblocks affects translation performance, while keeping document-level data constant.
The sentence-level portion is sourced from TowerBlocks \citep{TowerLLM} and uniformly sampled across languages.
Results shown in Figure~\ref{fig:sent_doc_prop} demonstrate that just $10\%$ sentence-level data is sufficient to maintain sentence-level quality, while higher proportions yield diminishing returns and reduce document-level performance. 
Allocating $90\%$ of training data to document-level pairs provides the best balance, enhancing document-level performance without compromising sentence-level capabilities.

\section{Related Work}

\paragraph{Document-Level Machine Translation.} Traditional sentence-level MT \citep{Sutskever_2014,bahdanau2014,Vaswani_2017} often leads to inconsistencies in terminology, tone, and context in longer texts \citep{DISCOURSE_STRUCT_1,DISCOURSE_STRUCT_2,GuoFeng_ST}. Document-level MT \citep{DocMT_Survey} improves coherence by incorporating broader context across sentences and paragraphs. Recent approaches include document embeddings \citep{DocEmbeddings-1,DocEmbeddings-2}, multi-encoder architectures \citep{ME-1,ME-2,ME-3}, and enhanced attention mechanisms for long-range dependencies \citep{ATTENTION-1,ATTENTION-2,ATTENTION-3}. 
While promising, progress is limited by the lack of high-quality document-level parallel data \citep{DocCorpora,GuoFeng_ST}, which hinders training in complex domains such as literary translation \citep{LITERARY-1,LITERARY-2}. 
Recent efforts have drawn on web crawls (e.g., ParaCrawl \citep{banon-etal-2020-paracrawl}, BWB \citep{BLONDE}), news data (e.g., WMT \citep{WMT24}), parliamentary records (e.g., Europarl \citep{koehn-2005-europarl}), public talks (e.g., IWSLT TED task \citep{IWSLT17}), literary texts (e.g., GuoFeng \citep{GuoFeng_ST}), and subtitles (e.g., OpenSubtitles \citep{OpenSub}), as well as large multilingual corpora (e.g., UNPC \citep{UNPC}). However, many of these suffer from domain constraints, noisy alignments, or unclear document boundaries.
We address these issues by curating \docblocks, a high-quality document-level dataset built from public sources, emphasizing clean alignment, coherent document structure, and careful filtering.

\paragraph{LLMs for Machine Translation.}
LLMs have recently demonstrated strong performance on MT tasks \citep{TowerLLM}. Their capacity to process long contexts opens new possibilities for modeling discourse structure and inter-sentence dependencies, which are essential for document-level translation.
To leverage these capabilities, new strategies are being explored. 
These include context-aware prompting, which uses relevant context directly in the prompt to guide translation \citep{DocMT_WANG,DocMT_WU}, and document-level fine-tuning, which adapts instruction-tuned models to achieve stronger performance \citep{DocMT_WU,wu-etal-2024-importance,li2024enhancingdocumentleveltranslationlarge}. 
In addition, agent-based approaches such as DELTA \citep{DELTA_LLM} and TRANSAGENTS \citep{AGENT_LLM} further improve consistency by maintaining structured memory across sentences, though they introduce substantial inference overhead.
Despite these advances, LLMs still face challenges in maintaining coherence over long documents due to limitations in training data \citep{wu-etal-2024-importance} and context modeling \citep{karpinska-iyyer-2023-large}.
To address these challenges, we fine-tune LLMs using \docblocks, enhancing their multilingual contextual understanding for document translation without incurring the high cost of full retraining.

\section{Conclusion}
This study advances the transition of state-of-the-art LLMs from sentence-level MT to document-level MT by addressing key challenges in contextual coherence and long-document consistency. Fine-tuning on the curated \docblocks dataset allows DocMT-LLMs to effectively manage long-range dependencies while preserving sentence-level translation quality. Moreover, context-aware prompt tuning and multi-resolution document-to-document training enhance adaptability across various translation paradigms, including document-to-document and several chunk-level methods. These findings provide a cost-effective and scalable approach to improving document-level translation quality.

\section*{Acknowledgments}
We thank the members of the SARDINE lab for their useful and constructive comments. This work was supported by the Portuguese Recovery and Resilience Plan through project C645008882-00000055 (Center for Responsible AI), by the EU’s Horizon Europe Research and Innovation Actions (UTTER, contract 101070631), by the project DECOLLAGE (ERC-2022-CoG 101088763), and by Fundação para a Ciência e Tecnologia through contract UIDB/50008/2020.

\bibliographystyle{colm2025_conference}
\bibliography{./bib.bib}

\begin{thebibliography}{76}
\providecommand{\natexlab}[1]{#1}
\providecommand{\url}[1]{\texttt{#1}}
\expandafter\ifx\csname urlstyle\endcsname\relax
  \providecommand{\doi}[1]{doi: #1}\else
  \providecommand{\doi}{doi: \begingroup \urlstyle{rm}\Url}\fi

\bibitem[{Alves} et~al.(2024){Alves}, {Pombal}, {Guerreiro}, {Martins}, {Alves}, {Farajian}, {Peters}, {Rei}, {Fernandes}, {Agrawal}, {Colombo}, {de Souza}, and {Martins}]{TowerLLM}
Duarte~M. {Alves}, Jos{\'e} {Pombal}, Nuno~M. {Guerreiro}, Pedro~H. {Martins}, Jo{\~a}o {Alves}, Amin {Farajian}, Ben {Peters}, Ricardo {Rei}, Patrick {Fernandes}, Sweta {Agrawal}, Pierre {Colombo}, Jos{\'e} G.~C. {de Souza}, and Andr{\'e} F.~T. {Martins}.
\newblock {Tower: An Open Multilingual Large Language Model for Translation-Related Tasks}.
\newblock \emph{arXiv e-prints}, art. arXiv:2402.17733, February 2024.
\newblock \doi{10.48550/arXiv.2402.17733}.

\bibitem[Anastasopoulos et~al.(2020)Anastasopoulos, Cattelan, Dou, Federico, Federmann, Genzel, Guzm{\'a}n, Hu, Hughes, Koehn, Lazar, Lewis, Neubig, Niu, {\"O}ktem, Paquin, Tang, and Tur]{anastasopoulos-etal-2020-tico}
Antonios Anastasopoulos, Alessandro Cattelan, Zi-Yi Dou, Marcello Federico, Christian Federmann, Dmitriy Genzel, Franscisco Guzm{\'a}n, Junjie Hu, Macduff Hughes, Philipp Koehn, Rosie Lazar, Will Lewis, Graham Neubig, Mengmeng Niu, Alp {\"O}ktem, Eric Paquin, Grace Tang, and Sylwia Tur.
\newblock {TICO}-19: the translation initiative for {CO}vid-19.
\newblock In Karin Verspoor, Kevin~Bretonnel Cohen, Michael Conway, Berry de~Bruijn, Mark Dredze, Rada Mihalcea, and Byron Wallace (eds.), \emph{Proceedings of the 1st Workshop on {NLP} for {COVID}-19 (Part 2) at {EMNLP} 2020}, Online, December 2020. Association for Computational Linguistics.
\newblock \doi{10.18653/v1/2020.nlpcovid19-2.5}.
\newblock URL \url{https://aclanthology.org/2020.nlpcovid19-2.5}.

\bibitem[{Bahdanau} et~al.(2014){Bahdanau}, {Cho}, and {Bengio}]{bahdanau2014}
Dzmitry {Bahdanau}, Kyunghyun {Cho}, and Yoshua {Bengio}.
\newblock {Neural Machine Translation by Jointly Learning to Align and Translate}.
\newblock \emph{arXiv e-prints}, art. arXiv:1409.0473, September 2014.
\newblock \doi{10.48550/arXiv.1409.0473}.

\bibitem[{Bai} et~al.(2023){Bai}, {Bai}, {Chu}, {Cui}, {Dang}, {Deng}, {Fan}, {Ge}, {Han}, {Huang}, {Hui}, {Ji}, {Li}, {Lin}, {Lin}, {Liu}, {Liu}, {Lu}, {Lu}, {Ma}, {Men}, {Ren}, {Ren}, {Tan}, {Tan}, {Tu}, {Wang}, {Wang}, {Wang}, {Wu}, {Xu}, {Xu}, {Yang}, {Yang}, {Yang}, {Yang}, {Yao}, {Yu}, {Yuan}, {Yuan}, {Zhang}, {Zhang}, {Zhang}, {Zhang}, {Zhou}, {Zhou}, {Zhou}, and {Zhu}]{QWEN_LLM}
Jinze {Bai}, Shuai {Bai}, Yunfei {Chu}, Zeyu {Cui}, Kai {Dang}, Xiaodong {Deng}, Yang {Fan}, Wenbin {Ge}, Yu~{Han}, Fei {Huang}, Binyuan {Hui}, Luo {Ji}, Mei {Li}, Junyang {Lin}, Runji {Lin}, Dayiheng {Liu}, Gao {Liu}, Chengqiang {Lu}, Keming {Lu}, Jianxin {Ma}, Rui {Men}, Xingzhang {Ren}, Xuancheng {Ren}, Chuanqi {Tan}, Sinan {Tan}, Jianhong {Tu}, Peng {Wang}, Shijie {Wang}, Wei {Wang}, Shengguang {Wu}, Benfeng {Xu}, Jin {Xu}, An~{Yang}, Hao {Yang}, Jian {Yang}, Shusheng {Yang}, Yang {Yao}, Bowen {Yu}, Hongyi {Yuan}, Zheng {Yuan}, Jianwei {Zhang}, Xingxuan {Zhang}, Yichang {Zhang}, Zhenru {Zhang}, Chang {Zhou}, Jingren {Zhou}, Xiaohuan {Zhou}, and Tianhang {Zhu}.
\newblock {Qwen Technical Report}.
\newblock \emph{arXiv e-prints}, art. arXiv:2309.16609, September 2023.
\newblock \doi{10.48550/arXiv.2309.16609}.

\bibitem[Ba{\~n}{\'o}n et~al.(2020)Ba{\~n}{\'o}n, Chen, Haddow, Heafield, Hoang, Espl{\`a}-Gomis, Forcada, Kamran, Kirefu, Koehn, Ortiz~Rojas, Pla~Sempere, Ram{\'i}rez-S{\'a}nchez, Sarr{\'i}as, Strelec, Thompson, Waites, Wiggins, and Zaragoza]{banon-etal-2020-paracrawl}
Marta Ba{\~n}{\'o}n, Pinzhen Chen, Barry Haddow, Kenneth Heafield, Hieu Hoang, Miquel Espl{\`a}-Gomis, Mikel~L. Forcada, Amir Kamran, Faheem Kirefu, Philipp Koehn, Sergio Ortiz~Rojas, Leopoldo Pla~Sempere, Gema Ram{\'i}rez-S{\'a}nchez, Elsa Sarr{\'i}as, Marek Strelec, Brian Thompson, William Waites, Dion Wiggins, and Jaume Zaragoza.
\newblock {P}ara{C}rawl: Web-scale acquisition of parallel corpora.
\newblock In Dan Jurafsky, Joyce Chai, Natalie Schluter, and Joel Tetreault (eds.), \emph{Proceedings of the 58th Annual Meeting of the Association for Computational Linguistics}, pp.\  4555--4567, Online, July 2020. Association for Computational Linguistics.
\newblock \doi{10.18653/v1/2020.acl-main.417}.
\newblock URL \url{https://aclanthology.org/2020.acl-main.417/}.

\bibitem[Bawden et~al.(2018)Bawden, Sennrich, Birch, and Haddow]{ME-3}
Rachel Bawden, Rico Sennrich, Alexandra Birch, and Barry Haddow.
\newblock Evaluating discourse phenomena in neural machine translation.
\newblock In Marilyn Walker, Heng Ji, and Amanda Stent (eds.), \emph{Proceedings of the 2018 Conference of the North {A}merican Chapter of the Association for Computational Linguistics: Human Language Technologies, Volume 1 (Long Papers)}, pp.\  1304--1313, New Orleans, Louisiana, June 2018. Association for Computational Linguistics.
\newblock \doi{10.18653/v1/N18-1118}.
\newblock URL \url{https://aclanthology.org/N18-1118}.

\bibitem[Castilho et~al.(2020)Castilho, Popovi{\'c}, and Way]{CASTILHO_SPAN}
Sheila Castilho, Maja Popovi{\'c}, and Andy Way.
\newblock On context span needed for machine translation evaluation.
\newblock In Nicoletta Calzolari, Fr{\'e}d{\'e}ric B{\'e}chet, Philippe Blache, Khalid Choukri, Christopher Cieri, Thierry Declerck, Sara Goggi, Hitoshi Isahara, Bente Maegaard, Joseph Mariani, H{\'e}l{\`e}ne Mazo, Asuncion Moreno, Jan Odijk, and Stelios Piperidis (eds.), \emph{Proceedings of the Twelfth Language Resources and Evaluation Conference}, pp.\  3735--3742, Marseille, France, May 2020. European Language Resources Association.
\newblock ISBN 979-10-95546-34-4.
\newblock URL \url{https://aclanthology.org/2020.lrec-1.461}.

\bibitem[Cettolo et~al.(2017)Cettolo, Federico, Bentivogli, Niehues, St{\"u}ker, Sudoh, Yoshino, and Federmann]{IWSLT17}
Mauro Cettolo, Marcello Federico, Luisa Bentivogli, Jan Niehues, Sebastian St{\"u}ker, Katsuhito Sudoh, Koichiro Yoshino, and Christian Federmann.
\newblock Overview of the {IWSLT} 2017 evaluation campaign.
\newblock In \emph{Proceedings of the 14th International Conference on Spoken Language Translation}, pp.\  2--14, Tokyo, Japan, December 14-15 2017. International Workshop on Spoken Language Translation.
\newblock URL \url{https://aclanthology.org/2017.iwslt-1.1}.

\bibitem[{Deutsch} et~al.(2025){Deutsch}, {Briakou}, {Caswell}, {Finkelstein}, {Galor}, {Juraska}, {Kovacs}, {Lui}, {Rei}, {Riesa}, {Rijhwani}, {Riley}, {Salesky}, {Trabelsi}, {Winkler}, {Zhang}, and {Freitag}]{WMT24++}
Daniel {Deutsch}, Eleftheria {Briakou}, Isaac {Caswell}, Mara {Finkelstein}, Rebecca {Galor}, Juraj {Juraska}, Geza {Kovacs}, Alison {Lui}, Ricardo {Rei}, Jason {Riesa}, Shruti {Rijhwani}, Parker {Riley}, Elizabeth {Salesky}, Firas {Trabelsi}, Stephanie {Winkler}, Biao {Zhang}, and Markus {Freitag}.
\newblock {WMT24++: Expanding the Language Coverage of WMT24 to 55 Languages \& Dialects}.
\newblock \emph{arXiv e-prints}, art. arXiv:2502.12404, February 2025.
\newblock \doi{10.48550/arXiv.2502.12404}.

\bibitem[Eikema \& Aziz(2022)Eikema and Aziz]{eikema-aziz-2022-sampling}
Bryan Eikema and Wilker Aziz.
\newblock Sampling-based approximations to minimum {B}ayes risk decoding for neural machine translation.
\newblock In Yoav Goldberg, Zornitsa Kozareva, and Yue Zhang (eds.), \emph{Proceedings of the 2022 Conference on Empirical Methods in Natural Language Processing}, pp.\  10978--10993, Abu Dhabi, United Arab Emirates, December 2022. Association for Computational Linguistics.
\newblock \doi{10.18653/v1/2022.emnlp-main.754}.
\newblock URL \url{https://aclanthology.org/2022.emnlp-main.754}.

\bibitem[Fernandes et~al.(2021)Fernandes, Yin, Neubig, and Martins]{FERNANDES_MEASURING}
Patrick Fernandes, Kayo Yin, Graham Neubig, and Andr{\'e} F.~T. Martins.
\newblock Measuring and increasing context usage in context-aware machine translation.
\newblock In Chengqing Zong, Fei Xia, Wenjie Li, and Roberto Navigli (eds.), \emph{Proceedings of the 59th Annual Meeting of the Association for Computational Linguistics and the 11th International Joint Conference on Natural Language Processing (Volume 1: Long Papers)}, pp.\  6467--6478, Online, August 2021. Association for Computational Linguistics.
\newblock \doi{10.18653/v1/2021.acl-long.505}.
\newblock URL \url{https://aclanthology.org/2021.acl-long.505}.

\bibitem[Freitag et~al.(2021)Freitag, Foster, Grangier, Ratnakar, Tan, and Macherey]{FREITAG_EEC}
Markus Freitag, George Foster, David Grangier, Viresh Ratnakar, Qijun Tan, and Wolfgang Macherey.
\newblock Experts, errors, and context: A large-scale study of human evaluation for machine translation.
\newblock \emph{Transactions of the Association for Computational Linguistics}, 9:\penalty0 1460--1474, 2021.
\newblock \doi{10.1162/tacl_a_00437}.
\newblock URL \url{https://aclanthology.org/2021.tacl-1.87}.

\bibitem[Ghazvininejad et~al.(2018)Ghazvininejad, Choi, and Knight]{LITERARY-1}
Marjan Ghazvininejad, Yejin Choi, and Kevin Knight.
\newblock Neural poetry translation.
\newblock In Marilyn Walker, Heng Ji, and Amanda Stent (eds.), \emph{Proceedings of the 2018 Conference of the North {A}merican Chapter of the Association for Computational Linguistics: Human Language Technologies, Volume 2 (Short Papers)}, pp.\  67--71, New Orleans, Louisiana, June 2018. Association for Computational Linguistics.
\newblock \doi{10.18653/v1/N18-2011}.
\newblock URL \url{https://aclanthology.org/N18-2011}.

\bibitem[Gong et~al.(2015)Gong, Zhang, and Zhou]{DocMetric-1}
Zhengxian Gong, Min Zhang, and Guodong Zhou.
\newblock Document-level machine translation evaluation with gist consistency and text cohesion.
\newblock In Bonnie Webber, Marine Carpuat, Andrei Popescu-Belis, and Christian Hardmeier (eds.), \emph{Proceedings of the Second Workshop on Discourse in Machine Translation}, pp.\  33--40, Lisbon, Portugal, September 2015. Association for Computational Linguistics.
\newblock \doi{10.18653/v1/W15-2504}.
\newblock URL \url{https://aclanthology.org/W15-2504}.

\bibitem[Grattafiori et~al.(2024)Grattafiori, Dubey, Jauhri, Pandey, Kadian, Al-Dahle, Letman, Mathur, Schelten, Vaughan, Yang, Fan, Goyal, Hartshorn, Yang, Mitra, Sravankumar, Korenev, Hinsvark, Rao, Zhang, Rodriguez, Gregerson, Spataru, Roziere, Biron, Tang, Chern, Caucheteux, Nayak, Bi, Marra, McConnell, Keller, Touret, Wu, Wong, Ferrer, Nikolaidis, Allonsius, Song, Pintz, Livshits, Wyatt, Esiobu, Choudhary, Mahajan, Garcia-Olano, Perino, Hupkes, Lakomkin, AlBadawy, Lobanova, Dinan, Smith, Radenovic, Guzmán, Zhang, Synnaeve, Lee, Anderson, Thattai, Nail, Mialon, Pang, Cucurell, Nguyen, Korevaar, Xu, Touvron, Zarov, Ibarra, Kloumann, Misra, Evtimov, Zhang, Copet, Lee, Geffert, Vranes, Park, Mahadeokar, Shah, van~der Linde, Billock, Hong, Lee, Fu, Chi, Huang, Liu, Wang, Yu, Bitton, Spisak, Park, Rocca, Johnstun, Saxe, Jia, Alwala, Prasad, Upasani, Plawiak, Li, Heafield, Stone, El-Arini, Iyer, Malik, Chiu, Bhalla, Lakhotia, Rantala-Yeary, van~der Maaten, Chen, Tan, Jenkins, Martin, Madaan, Malo, Blecher,
  Landzaat, de~Oliveira, Muzzi, Pasupuleti, Singh, Paluri, Kardas, Tsimpoukelli, Oldham, Rita, Pavlova, Kambadur, Lewis, Si, Singh, Hassan, Goyal, Torabi, Bashlykov, Bogoychev, Chatterji, Zhang, Duchenne, Çelebi, Alrassy, Zhang, Li, Vasic, Weng, Bhargava, Dubal, Krishnan, Koura, Xu, He, Dong, Srinivasan, Ganapathy, Calderer, Cabral, Stojnic, Raileanu, Maheswari, Girdhar, Patel, Sauvestre, Polidoro, Sumbaly, Taylor, Silva, Hou, Wang, Hosseini, Chennabasappa, Singh, Bell, Kim, Edunov, Nie, Narang, Raparthy, Shen, Wan, Bhosale, Zhang, Vandenhende, Batra, Whitman, Sootla, Collot, Gururangan, Borodinsky, Herman, Fowler, Sheasha, Georgiou, Scialom, Speckbacher, Mihaylov, Xiao, Karn, Goswami, Gupta, Ramanathan, Kerkez, Gonguet, Do, Vogeti, Albiero, Petrovic, Chu, Xiong, Fu, Meers, Martinet, Wang, Wang, Tan, Xia, Xie, Jia, Wang, Goldschlag, Gaur, Babaei, Wen, Song, Zhang, Li, Mao, Coudert, Yan, Chen, Papakipos, Singh, Srivastava, Jain, Kelsey, Shajnfeld, Gangidi, Victoria, Goldstand, Menon, Sharma, Boesenberg,
  Baevski, Feinstein, Kallet, Sangani, Teo, Yunus, Lupu, Alvarado, Caples, Gu, Ho, Poulton, Ryan, Ramchandani, Dong, Franco, Goyal, Saraf, Chowdhury, Gabriel, Bharambe, Eisenman, Yazdan, James, Maurer, Leonhardi, Huang, Loyd, Paola, Paranjape, Liu, Wu, Ni, Hancock, Wasti, Spence, Stojkovic, Gamido, Montalvo, Parker, Burton, Mejia, Liu, Wang, Kim, Zhou, Hu, Chu, Cai, Tindal, Feichtenhofer, Gao, Civin, Beaty, Kreymer, Li, Adkins, Xu, Testuggine, David, Parikh, Liskovich, Foss, Wang, Le, Holland, Dowling, Jamil, Montgomery, Presani, Hahn, Wood, Le, Brinkman, Arcaute, Dunbar, Smothers, Sun, Kreuk, Tian, Kokkinos, Ozgenel, Caggioni, Kanayet, Seide, Florez, Schwarz, Badeer, Swee, Halpern, Herman, Sizov, Guangyi, Zhang, Lakshminarayanan, Inan, Shojanazeri, Zou, Wang, Zha, Habeeb, Rudolph, Suk, Aspegren, Goldman, Zhan, Damlaj, Molybog, Tufanov, Leontiadis, Veliche, Gat, Weissman, Geboski, Kohli, Lam, Asher, Gaya, Marcus, Tang, Chan, Zhen, Reizenstein, Teboul, Zhong, Jin, Yang, Cummings, Carvill, Shepard, McPhie,
  Torres, Ginsburg, Wang, Wu, U, Saxena, Khandelwal, Zand, Matosich, Veeraraghavan, Michelena, Li, Jagadeesh, Huang, Chawla, Huang, Chen, Garg, A, Silva, Bell, Zhang, Guo, Yu, Moshkovich, Wehrstedt, Khabsa, Avalani, Bhatt, Mankus, Hasson, Lennie, Reso, Groshev, Naumov, Lathi, Keneally, Liu, Seltzer, Valko, Restrepo, Patel, Vyatskov, Samvelyan, Clark, Macey, Wang, Hermoso, Metanat, Rastegari, Bansal, Santhanam, Parks, White, Bawa, Singhal, Egebo, Usunier, Mehta, Laptev, Dong, Cheng, Chernoguz, Hart, Salpekar, Kalinli, Kent, Parekh, Saab, Balaji, Rittner, Bontrager, Roux, Dollar, Zvyagina, Ratanchandani, Yuvraj, Liang, Alao, Rodriguez, Ayub, Murthy, Nayani, Mitra, Parthasarathy, Li, Hogan, Battey, Wang, Howes, Rinott, Mehta, Siby, Bondu, Datta, Chugh, Hunt, Dhillon, Sidorov, Pan, Mahajan, Verma, Yamamoto, Ramaswamy, Lindsay, Lindsay, Feng, Lin, Zha, Patil, Shankar, Zhang, Zhang, Wang, Agarwal, Sajuyigbe, Chintala, Max, Chen, Kehoe, Satterfield, Govindaprasad, Gupta, Deng, Cho, Virk, Subramanian, Choudhury,
  Goldman, Remez, Glaser, Best, Koehler, Robinson, Li, Zhang, Matthews, Chou, Shaked, Vontimitta, Ajayi, Montanez, Mohan, Kumar, Mangla, Ionescu, Poenaru, Mihailescu, Ivanov, Li, Wang, Jiang, Bouaziz, Constable, Tang, Wu, Wang, Wu, Gao, Kleinman, Chen, Hu, Jia, Qi, Li, Zhang, Zhang, Adi, Nam, Yu, Wang, Zhao, Hao, Qian, Li, He, Rait, DeVito, Rosnbrick, Wen, Yang, Zhao, and Ma]{grattafiori2024llama3herdmodels}
Aaron Grattafiori, Abhimanyu Dubey, Abhinav Jauhri, Abhinav Pandey, Abhishek Kadian, Ahmad Al-Dahle, Aiesha Letman, Akhil Mathur, Alan Schelten, Alex Vaughan, Amy Yang, Angela Fan, Anirudh Goyal, Anthony Hartshorn, Aobo Yang, Archi Mitra, Archie Sravankumar, Artem Korenev, Arthur Hinsvark, Arun Rao, Aston Zhang, Aurelien Rodriguez, Austen Gregerson, Ava Spataru, Baptiste Roziere, Bethany Biron, Binh Tang, Bobbie Chern, Charlotte Caucheteux, Chaya Nayak, Chloe Bi, Chris Marra, Chris McConnell, Christian Keller, Christophe Touret, Chunyang Wu, Corinne Wong, Cristian~Canton Ferrer, Cyrus Nikolaidis, Damien Allonsius, Daniel Song, Danielle Pintz, Danny Livshits, Danny Wyatt, David Esiobu, Dhruv Choudhary, Dhruv Mahajan, Diego Garcia-Olano, Diego Perino, Dieuwke Hupkes, Egor Lakomkin, Ehab AlBadawy, Elina Lobanova, Emily Dinan, Eric~Michael Smith, Filip Radenovic, Francisco Guzmán, Frank Zhang, Gabriel Synnaeve, Gabrielle Lee, Georgia~Lewis Anderson, Govind Thattai, Graeme Nail, Gregoire Mialon, Guan Pang,
  Guillem Cucurell, Hailey Nguyen, Hannah Korevaar, Hu~Xu, Hugo Touvron, Iliyan Zarov, Imanol~Arrieta Ibarra, Isabel Kloumann, Ishan Misra, Ivan Evtimov, Jack Zhang, Jade Copet, Jaewon Lee, Jan Geffert, Jana Vranes, Jason Park, Jay Mahadeokar, Jeet Shah, Jelmer van~der Linde, Jennifer Billock, Jenny Hong, Jenya Lee, Jeremy Fu, Jianfeng Chi, Jianyu Huang, Jiawen Liu, Jie Wang, Jiecao Yu, Joanna Bitton, Joe Spisak, Jongsoo Park, Joseph Rocca, Joshua Johnstun, Joshua Saxe, Junteng Jia, Kalyan~Vasuden Alwala, Karthik Prasad, Kartikeya Upasani, Kate Plawiak, Ke~Li, Kenneth Heafield, Kevin Stone, Khalid El-Arini, Krithika Iyer, Kshitiz Malik, Kuenley Chiu, Kunal Bhalla, Kushal Lakhotia, Lauren Rantala-Yeary, Laurens van~der Maaten, Lawrence Chen, Liang Tan, Liz Jenkins, Louis Martin, Lovish Madaan, Lubo Malo, Lukas Blecher, Lukas Landzaat, Luke de~Oliveira, Madeline Muzzi, Mahesh Pasupuleti, Mannat Singh, Manohar Paluri, Marcin Kardas, Maria Tsimpoukelli, Mathew Oldham, Mathieu Rita, Maya Pavlova, Melanie Kambadur,
  Mike Lewis, Min Si, Mitesh~Kumar Singh, Mona Hassan, Naman Goyal, Narjes Torabi, Nikolay Bashlykov, Nikolay Bogoychev, Niladri Chatterji, Ning Zhang, Olivier Duchenne, Onur Çelebi, Patrick Alrassy, Pengchuan Zhang, Pengwei Li, Petar Vasic, Peter Weng, Prajjwal Bhargava, Pratik Dubal, Praveen Krishnan, Punit~Singh Koura, Puxin Xu, Qing He, Qingxiao Dong, Ragavan Srinivasan, Raj Ganapathy, Ramon Calderer, Ricardo~Silveira Cabral, Robert Stojnic, Roberta Raileanu, Rohan Maheswari, Rohit Girdhar, Rohit Patel, Romain Sauvestre, Ronnie Polidoro, Roshan Sumbaly, Ross Taylor, Ruan Silva, Rui Hou, Rui Wang, Saghar Hosseini, Sahana Chennabasappa, Sanjay Singh, Sean Bell, Seohyun~Sonia Kim, Sergey Edunov, Shaoliang Nie, Sharan Narang, Sharath Raparthy, Sheng Shen, Shengye Wan, Shruti Bhosale, Shun Zhang, Simon Vandenhende, Soumya Batra, Spencer Whitman, Sten Sootla, Stephane Collot, Suchin Gururangan, Sydney Borodinsky, Tamar Herman, Tara Fowler, Tarek Sheasha, Thomas Georgiou, Thomas Scialom, Tobias Speckbacher,
  Todor Mihaylov, Tong Xiao, Ujjwal Karn, Vedanuj Goswami, Vibhor Gupta, Vignesh Ramanathan, Viktor Kerkez, Vincent Gonguet, Virginie Do, Vish Vogeti, Vítor Albiero, Vladan Petrovic, Weiwei Chu, Wenhan Xiong, Wenyin Fu, Whitney Meers, Xavier Martinet, Xiaodong Wang, Xiaofang Wang, Xiaoqing~Ellen Tan, Xide Xia, Xinfeng Xie, Xuchao Jia, Xuewei Wang, Yaelle Goldschlag, Yashesh Gaur, Yasmine Babaei, Yi~Wen, Yiwen Song, Yuchen Zhang, Yue Li, Yuning Mao, Zacharie~Delpierre Coudert, Zheng Yan, Zhengxing Chen, Zoe Papakipos, Aaditya Singh, Aayushi Srivastava, Abha Jain, Adam Kelsey, Adam Shajnfeld, Adithya Gangidi, Adolfo Victoria, Ahuva Goldstand, Ajay Menon, Ajay Sharma, Alex Boesenberg, Alexei Baevski, Allie Feinstein, Amanda Kallet, Amit Sangani, Amos Teo, Anam Yunus, Andrei Lupu, Andres Alvarado, Andrew Caples, Andrew Gu, Andrew Ho, Andrew Poulton, Andrew Ryan, Ankit Ramchandani, Annie Dong, Annie Franco, Anuj Goyal, Aparajita Saraf, Arkabandhu Chowdhury, Ashley Gabriel, Ashwin Bharambe, Assaf Eisenman, Azadeh
  Yazdan, Beau James, Ben Maurer, Benjamin Leonhardi, Bernie Huang, Beth Loyd, Beto~De Paola, Bhargavi Paranjape, Bing Liu, Bo~Wu, Boyu Ni, Braden Hancock, Bram Wasti, Brandon Spence, Brani Stojkovic, Brian Gamido, Britt Montalvo, Carl Parker, Carly Burton, Catalina Mejia, Ce~Liu, Changhan Wang, Changkyu Kim, Chao Zhou, Chester Hu, Ching-Hsiang Chu, Chris Cai, Chris Tindal, Christoph Feichtenhofer, Cynthia Gao, Damon Civin, Dana Beaty, Daniel Kreymer, Daniel Li, David Adkins, David Xu, Davide Testuggine, Delia David, Devi Parikh, Diana Liskovich, Didem Foss, Dingkang Wang, Duc Le, Dustin Holland, Edward Dowling, Eissa Jamil, Elaine Montgomery, Eleonora Presani, Emily Hahn, Emily Wood, Eric-Tuan Le, Erik Brinkman, Esteban Arcaute, Evan Dunbar, Evan Smothers, Fei Sun, Felix Kreuk, Feng Tian, Filippos Kokkinos, Firat Ozgenel, Francesco Caggioni, Frank Kanayet, Frank Seide, Gabriela~Medina Florez, Gabriella Schwarz, Gada Badeer, Georgia Swee, Gil Halpern, Grant Herman, Grigory Sizov, Guangyi, Zhang, Guna
  Lakshminarayanan, Hakan Inan, Hamid Shojanazeri, Han Zou, Hannah Wang, Hanwen Zha, Haroun Habeeb, Harrison Rudolph, Helen Suk, Henry Aspegren, Hunter Goldman, Hongyuan Zhan, Ibrahim Damlaj, Igor Molybog, Igor Tufanov, Ilias Leontiadis, Irina-Elena Veliche, Itai Gat, Jake Weissman, James Geboski, James Kohli, Janice Lam, Japhet Asher, Jean-Baptiste Gaya, Jeff Marcus, Jeff Tang, Jennifer Chan, Jenny Zhen, Jeremy Reizenstein, Jeremy Teboul, Jessica Zhong, Jian Jin, Jingyi Yang, Joe Cummings, Jon Carvill, Jon Shepard, Jonathan McPhie, Jonathan Torres, Josh Ginsburg, Junjie Wang, Kai Wu, Kam~Hou U, Karan Saxena, Kartikay Khandelwal, Katayoun Zand, Kathy Matosich, Kaushik Veeraraghavan, Kelly Michelena, Keqian Li, Kiran Jagadeesh, Kun Huang, Kunal Chawla, Kyle Huang, Lailin Chen, Lakshya Garg, Lavender A, Leandro Silva, Lee Bell, Lei Zhang, Liangpeng Guo, Licheng Yu, Liron Moshkovich, Luca Wehrstedt, Madian Khabsa, Manav Avalani, Manish Bhatt, Martynas Mankus, Matan Hasson, Matthew Lennie, Matthias Reso, Maxim
  Groshev, Maxim Naumov, Maya Lathi, Meghan Keneally, Miao Liu, Michael~L. Seltzer, Michal Valko, Michelle Restrepo, Mihir Patel, Mik Vyatskov, Mikayel Samvelyan, Mike Clark, Mike Macey, Mike Wang, Miquel~Jubert Hermoso, Mo~Metanat, Mohammad Rastegari, Munish Bansal, Nandhini Santhanam, Natascha Parks, Natasha White, Navyata Bawa, Nayan Singhal, Nick Egebo, Nicolas Usunier, Nikhil Mehta, Nikolay~Pavlovich Laptev, Ning Dong, Norman Cheng, Oleg Chernoguz, Olivia Hart, Omkar Salpekar, Ozlem Kalinli, Parkin Kent, Parth Parekh, Paul Saab, Pavan Balaji, Pedro Rittner, Philip Bontrager, Pierre Roux, Piotr Dollar, Polina Zvyagina, Prashant Ratanchandani, Pritish Yuvraj, Qian Liang, Rachad Alao, Rachel Rodriguez, Rafi Ayub, Raghotham Murthy, Raghu Nayani, Rahul Mitra, Rangaprabhu Parthasarathy, Raymond Li, Rebekkah Hogan, Robin Battey, Rocky Wang, Russ Howes, Ruty Rinott, Sachin Mehta, Sachin Siby, Sai~Jayesh Bondu, Samyak Datta, Sara Chugh, Sara Hunt, Sargun Dhillon, Sasha Sidorov, Satadru Pan, Saurabh Mahajan,
  Saurabh Verma, Seiji Yamamoto, Sharadh Ramaswamy, Shaun Lindsay, Shaun Lindsay, Sheng Feng, Shenghao Lin, Shengxin~Cindy Zha, Shishir Patil, Shiva Shankar, Shuqiang Zhang, Shuqiang Zhang, Sinong Wang, Sneha Agarwal, Soji Sajuyigbe, Soumith Chintala, Stephanie Max, Stephen Chen, Steve Kehoe, Steve Satterfield, Sudarshan Govindaprasad, Sumit Gupta, Summer Deng, Sungmin Cho, Sunny Virk, Suraj Subramanian, Sy~Choudhury, Sydney Goldman, Tal Remez, Tamar Glaser, Tamara Best, Thilo Koehler, Thomas Robinson, Tianhe Li, Tianjun Zhang, Tim Matthews, Timothy Chou, Tzook Shaked, Varun Vontimitta, Victoria Ajayi, Victoria Montanez, Vijai Mohan, Vinay~Satish Kumar, Vishal Mangla, Vlad Ionescu, Vlad Poenaru, Vlad~Tiberiu Mihailescu, Vladimir Ivanov, Wei Li, Wenchen Wang, Wenwen Jiang, Wes Bouaziz, Will Constable, Xiaocheng Tang, Xiaojian Wu, Xiaolan Wang, Xilun Wu, Xinbo Gao, Yaniv Kleinman, Yanjun Chen, Ye~Hu, Ye~Jia, Ye~Qi, Yenda Li, Yilin Zhang, Ying Zhang, Yossi Adi, Youngjin Nam, Yu, Wang, Yu~Zhao, Yuchen Hao, Yundi
  Qian, Yunlu Li, Yuzi He, Zach Rait, Zachary DeVito, Zef Rosnbrick, Zhaoduo Wen, Zhenyu Yang, Zhiwei Zhao, and Zhiyu Ma.
\newblock The llama 3 herd of models, 2024.
\newblock URL \url{https://arxiv.org/abs/2407.21783}.

\bibitem[He et~al.(2024)He, Liang, Jiao, Zhang, Yang, Wang, Tu, Shi, and Wang]{LLM_HumanLike_Translation}
Zhiwei He, Tian Liang, Wenxiang Jiao, Zhuosheng Zhang, Yujiu Yang, Rui Wang, Zhaopeng Tu, Shuming Shi, and Xing Wang.
\newblock Exploring human-like translation strategy with large language models.
\newblock \emph{Transactions of the Association for Computational Linguistics}, 12:\penalty0 229--246, 2024.
\newblock \doi{10.1162/tacl_a_00642}.
\newblock URL \url{https://aclanthology.org/2024.tacl-1.13}.

\bibitem[{Hendy} et~al.(2023){Hendy}, {Abdelrehim}, {Sharaf}, {Raunak}, {Gabr}, {Matsushita}, {Kim}, {Afify}, and {Awadalla}]{GPT_MT}
Amr {Hendy}, Mohamed {Abdelrehim}, Amr {Sharaf}, Vikas {Raunak}, Mohamed {Gabr}, Hitokazu {Matsushita}, Young~Jin {Kim}, Mohamed {Afify}, and Hany~Hassan {Awadalla}.
\newblock {How Good Are GPT Models at Machine Translation? A Comprehensive Evaluation}.
\newblock \emph{arXiv e-prints}, art. arXiv:2302.09210, February 2023.
\newblock \doi{10.48550/arXiv.2302.09210}.

\bibitem[{Holtzman} et~al.(2019){Holtzman}, {Buys}, {Du}, {Forbes}, and {Choi}]{2019arXiv190409751H}
Ari {Holtzman}, Jan {Buys}, Li~{Du}, Maxwell {Forbes}, and Yejin {Choi}.
\newblock {The Curious Case of Neural Text Degeneration}.
\newblock \emph{arXiv e-prints}, art. arXiv:1904.09751, April 2019.
\newblock \doi{10.48550/arXiv.1904.09751}.

\bibitem[Hu \& Wan(2023)Hu and Wan]{DISCOURSE_STRUCT_2}
Xinyu Hu and Xiaojun Wan.
\newblock Exploring discourse structure in document-level machine translation.
\newblock In Houda Bouamor, Juan Pino, and Kalika Bali (eds.), \emph{Proceedings of the 2023 Conference on Empirical Methods in Natural Language Processing}, pp.\  13889--13902, Singapore, December 2023. Association for Computational Linguistics.
\newblock \doi{10.18653/v1/2023.emnlp-main.857}.
\newblock URL \url{https://aclanthology.org/2023.emnlp-main.857}.

\bibitem[Huo et~al.(2020)Huo, Herold, Gao, Dahlmann, Khadivi, and Ney]{DocEmbeddings-2}
Jingjing Huo, Christian Herold, Yingbo Gao, Leonard Dahlmann, Shahram Khadivi, and Hermann Ney.
\newblock Diving deep into context-aware neural machine translation.
\newblock In Lo{\"\i}c Barrault, Ond{\v{r}}ej Bojar, Fethi Bougares, Rajen Chatterjee, Marta~R. Costa-juss{\`a}, Christian Federmann, Mark Fishel, Alexander Fraser, Yvette Graham, Paco Guzman, Barry Haddow, Matthias Huck, Antonio~Jimeno Yepes, Philipp Koehn, Andr{\'e} Martins, Makoto Morishita, Christof Monz, Masaaki Nagata, Toshiaki Nakazawa, and Matteo Negri (eds.), \emph{Proceedings of the Fifth Conference on Machine Translation}, pp.\  604--616, Online, November 2020. Association for Computational Linguistics.
\newblock URL \url{https://aclanthology.org/2020.wmt-1.71}.

\bibitem[{Jiang} et~al.(2023){Jiang}, {Sablayrolles}, {Mensch}, {Bamford}, {Singh Chaplot}, {de las Casas}, {Bressand}, {Lengyel}, {Lample}, {Saulnier}, {Renard Lavaud}, {Lachaux}, {Stock}, {Le Scao}, {Lavril}, {Wang}, {Lacroix}, and {El Sayed}]{MISTRAL}
Albert~Q. {Jiang}, Alexandre {Sablayrolles}, Arthur {Mensch}, Chris {Bamford}, Devendra {Singh Chaplot}, Diego {de las Casas}, Florian {Bressand}, Gianna {Lengyel}, Guillaume {Lample}, Lucile {Saulnier}, L{\'e}lio {Renard Lavaud}, Marie-Anne {Lachaux}, Pierre {Stock}, Teven {Le Scao}, Thibaut {Lavril}, Thomas {Wang}, Timoth{\'e}e {Lacroix}, and William {El Sayed}.
\newblock {Mistral 7B}.
\newblock \emph{arXiv e-prints}, art. arXiv:2310.06825, October 2023.
\newblock \doi{10.48550/arXiv.2310.06825}.

\bibitem[Jiang et~al.(2022)Jiang, Liu, Ma, Zhang, Yang, Huang, Sennrich, Cotterell, Sachan, and Zhou]{BLONDE}
Yuchen Jiang, Tianyu Liu, Shuming Ma, Dongdong Zhang, Jian Yang, Haoyang Huang, Rico Sennrich, Ryan Cotterell, Mrinmaya Sachan, and Ming Zhou.
\newblock {BlonDe}: An automatic evaluation metric for document-level machine translation.
\newblock In Marine Carpuat, Marie-Catherine de~Marneffe, and Ivan~Vladimir Meza~Ruiz (eds.), \emph{Proceedings of the 2022 Conference of the North American Chapter of the Association for Computational Linguistics: Human Language Technologies}, pp.\  1550--1565, Seattle, United States, July 2022. Association for Computational Linguistics.
\newblock \doi{10.18653/v1/2022.naacl-main.111}.
\newblock URL \url{https://aclanthology.org/2022.naacl-main.111}.

\bibitem[Karpinska \& Iyyer(2023)Karpinska and Iyyer]{karpinska-iyyer-2023-large}
Marzena Karpinska and Mohit Iyyer.
\newblock Large language models effectively leverage document-level context for literary translation, but critical errors persist.
\newblock In Philipp Koehn, Barry Haddow, Tom Kocmi, and Christof Monz (eds.), \emph{Proceedings of the Eighth Conference on Machine Translation}, pp.\  419--451, Singapore, December 2023. Association for Computational Linguistics.
\newblock \doi{10.18653/v1/2023.wmt-1.41}.
\newblock URL \url{https://aclanthology.org/2023.wmt-1.41/}.

\bibitem[{Kingma} \& {Ba}(2014){Kingma} and {Ba}]{2014arXiv1412.6980K}
Diederik~P. {Kingma} and Jimmy {Ba}.
\newblock {Adam: A Method for Stochastic Optimization}.
\newblock \emph{arXiv e-prints}, art. arXiv:1412.6980, December 2014.
\newblock \doi{10.48550/arXiv.1412.6980}.

\bibitem[Kocmi et~al.(2023)Kocmi, Avramidis, Bawden, Bojar, Dvorkovich, Federmann, Fishel, Freitag, Gowda, Grundkiewicz, Haddow, Koehn, Marie, Monz, Morishita, Murray, Nagata, Nakazawa, Popel, Popovi{\'c}, and Shmatova]{kocmi-etal-2023-findings}
Tom Kocmi, Eleftherios Avramidis, Rachel Bawden, Ond{\v{r}}ej Bojar, Anton Dvorkovich, Christian Federmann, Mark Fishel, Markus Freitag, Thamme Gowda, Roman Grundkiewicz, Barry Haddow, Philipp Koehn, Benjamin Marie, Christof Monz, Makoto Morishita, Kenton Murray, Makoto Nagata, Toshiaki Nakazawa, Martin Popel, Maja Popovi{\'c}, and Mariya Shmatova.
\newblock Findings of the 2023 conference on machine translation ({WMT}23): {LLM}s are here but not quite there yet.
\newblock In Philipp Koehn, Barry Haddow, Tom Kocmi, and Christof Monz (eds.), \emph{Proceedings of the Eighth Conference on Machine Translation}, pp.\  1--42, Singapore, December 2023. Association for Computational Linguistics.
\newblock \doi{10.18653/v1/2023.wmt-1.1}.
\newblock URL \url{https://aclanthology.org/2023.wmt-1.1}.

\bibitem[Kocmi et~al.(2024)Kocmi, Avramidis, Bawden, Bojar, Dvorkovich, Federmann, Fishel, Freitag, Gowda, Grundkiewicz, Haddow, Karpinska, Koehn, Marie, Monz, Murray, Nagata, Popel, Popovi{\'c}, Shmatova, Steingr{\'i}msson, and Zouhar]{WMT24}
Tom Kocmi, Eleftherios Avramidis, Rachel Bawden, Ond{\v{r}}ej Bojar, Anton Dvorkovich, Christian Federmann, Mark Fishel, Markus Freitag, Thamme Gowda, Roman Grundkiewicz, Barry Haddow, Marzena Karpinska, Philipp Koehn, Benjamin Marie, Christof Monz, Kenton Murray, Masaaki Nagata, Martin Popel, Maja Popovi{\'c}, Mariya Shmatova, Steinth{\'o}r Steingr{\'i}msson, and Vil{\'e}m Zouhar.
\newblock Findings of the {WMT}24 general machine translation shared task: The {LLM} era is here but {MT} is not solved yet.
\newblock In Barry Haddow, Tom Kocmi, Philipp Koehn, and Christof Monz (eds.), \emph{Proceedings of the Ninth Conference on Machine Translation}, pp.\  1--46, Miami, Florida, USA, November 2024. Association for Computational Linguistics.
\newblock \doi{10.18653/v1/2024.wmt-1.1}.
\newblock URL \url{https://aclanthology.org/2024.wmt-1.1/}.

\bibitem[Koehn(2005)]{koehn-2005-europarl}
Philipp Koehn.
\newblock {E}uroparl: A parallel corpus for statistical machine translation.
\newblock In \emph{Proceedings of Machine Translation Summit X: Papers}, pp.\  79--86, Phuket, Thailand, September 13-15 2005.
\newblock URL \url{https://aclanthology.org/2005.mtsummit-papers.11/}.

\bibitem[Kumar \& Byrne(2004)Kumar and Byrne]{kumar-byrne-2004-minimum}
Shankar Kumar and William Byrne.
\newblock Minimum {B}ayes-risk decoding for statistical machine translation.
\newblock In \emph{Proceedings of the Human Language Technology Conference of the North {A}merican Chapter of the Association for Computational Linguistics: {HLT}-{NAACL} 2004}, pp.\  169--176, Boston, Massachusetts, USA, May 2 - May 7 2004. Association for Computational Linguistics.
\newblock URL \url{https://aclanthology.org/N04-1022}.

\bibitem[L{\"a}ubli et~al.(2018)L{\"a}ubli, Sennrich, and Volk]{LAUBLI}
Samuel L{\"a}ubli, Rico Sennrich, and Martin Volk.
\newblock Has machine translation achieved human parity? a case for document-level evaluation.
\newblock In Ellen Riloff, David Chiang, Julia Hockenmaier, and Jun{'}ichi Tsujii (eds.), \emph{Proceedings of the 2018 Conference on Empirical Methods in Natural Language Processing}, pp.\  4791--4796, Brussels, Belgium, October-November 2018. Association for Computational Linguistics.
\newblock \doi{10.18653/v1/D18-1512}.
\newblock URL \url{https://aclanthology.org/D18-1512}.

\bibitem[Li et~al.(2024)Li, Li, Jiang, and Zhang]{li2024enhancingdocumentleveltranslationlarge}
Yachao Li, Junhui Li, Jing Jiang, and Min Zhang.
\newblock Enhancing document-level translation of large language model via translation mixed-instructions, 2024.
\newblock URL \url{https://arxiv.org/abs/2401.08088}.

\bibitem[Lison \& Tiedemann(2016)Lison and Tiedemann]{OpenSub}
Pierre Lison and J{\"o}rg Tiedemann.
\newblock {O}pen{S}ubtitles2016: Extracting large parallel corpora from movie and {TV} subtitles.
\newblock In Nicoletta Calzolari, Khalid Choukri, Thierry Declerck, Sara Goggi, Marko Grobelnik, Bente Maegaard, Joseph Mariani, Helene Mazo, Asuncion Moreno, Jan Odijk, and Stelios Piperidis (eds.), \emph{Proceedings of the Tenth International Conference on Language Resources and Evaluation ({LREC}'16)}, pp.\  923--929, Portoro{\v{z}}, Slovenia, May 2016. European Language Resources Association (ELRA).
\newblock URL \url{https://aclanthology.org/L16-1147}.

\bibitem[Liu \& Zhang(2020)Liu and Zhang]{DocCorpora}
Siyou Liu and Xiaojun Zhang.
\newblock Corpora for document-level neural machine translation.
\newblock In Nicoletta Calzolari, Fr{\'e}d{\'e}ric B{\'e}chet, Philippe Blache, Khalid Choukri, Christopher Cieri, Thierry Declerck, Sara Goggi, Hitoshi Isahara, Bente Maegaard, Joseph Mariani, H{\'e}l{\`e}ne Mazo, Asuncion Moreno, Jan Odijk, and Stelios Piperidis (eds.), \emph{Proceedings of the Twelfth Language Resources and Evaluation Conference}, pp.\  3775--3781, Marseille, France, May 2020. European Language Resources Association.
\newblock ISBN 979-10-95546-34-4.
\newblock URL \url{https://aclanthology.org/2020.lrec-1.466}.

\bibitem[Liu et~al.(2020)Liu, Gu, Goyal, Li, Edunov, Ghazvininejad, Lewis, and Zettlemoyer]{liu-etal-2020-multilingual-denoising}
Yinhan Liu, Jiatao Gu, Naman Goyal, Xian Li, Sergey Edunov, Marjan Ghazvininejad, Mike Lewis, and Luke Zettlemoyer.
\newblock Multilingual denoising pre-training for neural machine translation.
\newblock \emph{Transactions of the Association for Computational Linguistics}, 8:\penalty0 726--742, 2020.
\newblock \doi{10.1162/tacl_a_00343}.
\newblock URL \url{https://aclanthology.org/2020.tacl-1.47}.

\bibitem[Lui \& Baldwin(2012)Lui and Baldwin]{lui-baldwin-2012-langid}
Marco Lui and Timothy Baldwin.
\newblock langid.py: An off-the-shelf language identification tool.
\newblock In Min Zhang (ed.), \emph{Proceedings of the {ACL} 2012 System Demonstrations}, pp.\  25--30, Jeju Island, Korea, July 2012. Association for Computational Linguistics.
\newblock URL \url{https://aclanthology.org/P12-3005}.

\bibitem[Lyu et~al.(2021)Lyu, Li, Gong, and Zhang]{lyu-etal-2021-encouraging}
Xinglin Lyu, Junhui Li, Zhengxian Gong, and Min Zhang.
\newblock Encouraging lexical translation consistency for document-level neural machine translation.
\newblock In Marie-Francine Moens, Xuanjing Huang, Lucia Specia, and Scott Wen-tau Yih (eds.), \emph{Proceedings of the 2021 Conference on Empirical Methods in Natural Language Processing}, pp.\  3265--3277, Online and Punta Cana, Dominican Republic, November 2021. Association for Computational Linguistics.
\newblock \doi{10.18653/v1/2021.emnlp-main.262}.
\newblock URL \url{https://aclanthology.org/2021.emnlp-main.262}.

\bibitem[Mac{\'e} \& Servan(2019)Mac{\'e} and Servan]{DocEmbeddings-1}
Valentin Mac{\'e} and Christophe Servan.
\newblock Using whole document context in neural machine translation.
\newblock In Jan Niehues, Rolando Cattoni, Sebastian St{\"u}ker, Matteo Negri, Marco Turchi, Thanh-Le Ha, Elizabeth Salesky, Ramon Sanabria, Loic Barrault, Lucia Specia, and Marcello Federico (eds.), \emph{Proceedings of the 16th International Conference on Spoken Language Translation}, Hong Kong, November 2-3 2019. Association for Computational Linguistics.
\newblock URL \url{https://aclanthology.org/2019.iwslt-1.21}.

\bibitem[{Martins} et~al.(2024){Martins}, {Fernandes}, {Alves}, {Guerreiro}, {Rei}, {Alves}, {Pombal}, {Farajian}, {Faysse}, {Klimaszewski}, {Colombo}, {Haddow}, {de Souza}, {Birch}, and {Martins}]{EuroLLM}
Pedro~Henrique {Martins}, Patrick {Fernandes}, Jo{\~a}o {Alves}, Nuno~M. {Guerreiro}, Ricardo {Rei}, Duarte~M. {Alves}, Jos{\'e} {Pombal}, Amin {Farajian}, Manuel {Faysse}, Mateusz {Klimaszewski}, Pierre {Colombo}, Barry {Haddow}, Jos{\'e} G.~C. {de Souza}, Alexandra {Birch}, and Andr{\'e} F.~T. {Martins}.
\newblock {EuroLLM: Multilingual Language Models for Europe}.
\newblock \emph{arXiv e-prints}, art. arXiv:2409.16235, September 2024.
\newblock \doi{10.48550/arXiv.2409.16235}.

\bibitem[Maruf et~al.(2021)Maruf, Saleh, and Haffari]{DocMT_Survey}
Sameen Maruf, Fahimeh Saleh, and Gholamreza Haffari.
\newblock A survey on document-level neural machine translation: Methods and evaluation.
\newblock \emph{ACM Comput. Surv.}, 54\penalty0 (2), March 2021.
\newblock ISSN 0360-0300.
\newblock \doi{10.1145/3441691}.
\newblock URL \url{https://doi.org/10.1145/3441691}.

\bibitem[Miculicich et~al.(2018)Miculicich, Ram, Pappas, and Henderson]{ATTENTION-2}
Lesly Miculicich, Dhananjay Ram, Nikolaos Pappas, and James Henderson.
\newblock Document-level neural machine translation with hierarchical attention networks.
\newblock In Ellen Riloff, David Chiang, Julia Hockenmaier, and Jun{'}ichi Tsujii (eds.), \emph{Proceedings of the 2018 Conference on Empirical Methods in Natural Language Processing}, pp.\  2947--2954, Brussels, Belgium, October-November 2018. Association for Computational Linguistics.
\newblock \doi{10.18653/v1/D18-1325}.
\newblock URL \url{https://aclanthology.org/D18-1325}.

\bibitem[{NLLB Team} et~al.(2022){NLLB Team}, {Costa-juss{\`a}}, {Cross}, {{\c{C}}elebi}, {Elbayad}, {Heafield}, {Heffernan}, {Kalbassi}, {Lam}, {Licht}, {Maillard}, {Sun}, {Wang}, {Wenzek}, {Youngblood}, {Akula}, {Barrault}, {Mejia Gonzalez}, {Hansanti}, {Hoffman}, {Jarrett}, {Sadagopan}, {Rowe}, {Spruit}, {Tran}, {Andrews}, {Fazil Ayan}, {Bhosale}, {Edunov}, {Fan}, {Gao}, {Goswami}, {Guzm{\'a}n}, {Koehn}, {Mourachko}, {Ropers}, {Saleem}, {Schwenk}, and {Wang}]{NLLB}
{NLLB Team}, Marta~R. {Costa-juss{\`a}}, James {Cross}, Onur {{\c{C}}elebi}, Maha {Elbayad}, Kenneth {Heafield}, Kevin {Heffernan}, Elahe {Kalbassi}, Janice {Lam}, Daniel {Licht}, Jean {Maillard}, Anna {Sun}, Skyler {Wang}, Guillaume {Wenzek}, Al~{Youngblood}, Bapi {Akula}, Loic {Barrault}, Gabriel {Mejia Gonzalez}, Prangthip {Hansanti}, John {Hoffman}, Semarley {Jarrett}, Kaushik~Ram {Sadagopan}, Dirk {Rowe}, Shannon {Spruit}, Chau {Tran}, Pierre {Andrews}, Necip {Fazil Ayan}, Shruti {Bhosale}, Sergey {Edunov}, Angela {Fan}, Cynthia {Gao}, Vedanuj {Goswami}, Francisco {Guzm{\'a}n}, Philipp {Koehn}, Alexandre {Mourachko}, Christophe {Ropers}, Safiyyah {Saleem}, Holger {Schwenk}, and Jeff {Wang}.
\newblock {No Language Left Behind: Scaling Human-Centered Machine Translation}.
\newblock \emph{arXiv e-prints}, art. arXiv:2207.04672, July 2022.
\newblock \doi{10.48550/arXiv.2207.04672}.

\bibitem[{OpenAI}(2023)]{chatml}
{OpenAI}, 2023.
\newblock URL \url{https://github.com/openai/openai-python/blob/release-v0.28.1/chatml.md}.

\bibitem[{OpenAI} et~al.(2023){OpenAI}, {Achiam}, {Adler}, {Agarwal}, {Ahmad}, {Akkaya}, {Leoni Aleman}, {Almeida}, {Altenschmidt}, {Altman}, {Anadkat}, {Avila}, {Babuschkin}, {Balaji}, {Balcom}, {Baltescu}, {Bao}, {Bavarian}, {Belgum}, {Bello}, {Berdine}, {Bernadett-Shapiro}, {Berner}, {Bogdonoff}, {Boiko}, {Boyd}, {Brakman}, {Brockman}, {Brooks}, {Brundage}, {Button}, {Cai}, {Campbell}, {Cann}, {Carey}, {Carlson}, {Carmichael}, {Chan}, {Chang}, {Chantzis}, {Chen}, {Chen}, {Chen}, {Chen}, {Chen}, {Chess}, {Cho}, {Chu}, {Chung}, {Cummings}, {Currier}, {Dai}, {Decareaux}, {Degry}, {Deutsch}, {Deville}, {Dhar}, {Dohan}, {Dowling}, {Dunning}, {Ecoffet}, {Eleti}, {Eloundou}, {Farhi}, {Fedus}, {Felix}, {Posada Fishman}, {Forte}, {Fulford}, {Gao}, {Georges}, {Gibson}, {Goel}, {Gogineni}, {Goh}, {Gontijo-Lopes}, {Gordon}, {Grafstein}, {Gray}, {Greene}, {Gross}, {Gu}, {Guo}, {Hallacy}, {Han}, {Harris}, {He}, {Heaton}, {Heidecke}, {Hesse}, {Hickey}, {Hickey}, {Hoeschele}, {Houghton}, {Hsu}, {Hu}, {Hu}, {Huizinga},
  {Jain}, {Jain}, {Jang}, {Jiang}, {Jiang}, {Jin}, {Jin}, {Jomoto}, {Jonn}, {Jun}, {Kaftan}, {Kaiser}, {Kamali}, {Kanitscheider}, {Shirish Keskar}, {Khan}, {Kilpatrick}, {Kim}, {Kim}, {Kim}, {Hendrik Kirchner}, {Kiros}, {Knight}, {Kokotajlo}, {Kondraciuk}, {Kondrich}, {Konstantinidis}, {Kosic}, {Krueger}, {Kuo}, {Lampe}, {Lan}, {Lee}, {Leike}, {Leung}, {Levy}, {Li}, {Lim}, {Lin}, {Lin}, {Litwin}, {Lopez}, {Lowe}, {Lue}, {Makanju}, {Malfacini}, {Manning}, {Markov}, {Markovski}, {Martin}, {Mayer}, {Mayne}, {McGrew}, {McKinney}, {McLeavey}, {McMillan}, {McNeil}, {Medina}, {Mehta}, {Menick}, {Metz}, {Mishchenko}, {Mishkin}, {Monaco}, {Morikawa}, {Mossing}, {Mu}, {Murati}, {Murk}, {M{\'e}ly}, {Nair}, {Nakano}, {Nayak}, {Neelakantan}, {Ngo}, {Noh}, {Ouyang}, {O'Keefe}, {Pachocki}, {Paino}, {Palermo}, {Pantuliano}, {Parascandolo}, {Parish}, {Parparita}, {Passos}, {Pavlov}, {Peng}, {Perelman}, {de Avila Belbute Peres}, {Petrov}, {Ponde de Oliveira Pinto}, {Michael}, {Pokorny}, {Pokrass}, {Pong}, {Powell}, {Power},
  {Power}, {Proehl}, {Puri}, {Radford}, {Rae}, {Ramesh}, {Raymond}, {Real}, {Rimbach}, {Ross}, {Rotsted}, {Roussez}, {Ryder}, {Saltarelli}, {Sanders}, {Santurkar}, {Sastry}, {Schmidt}, {Schnurr}, {Schulman}, {Selsam}, {Sheppard}, {Sherbakov}, {Shieh}, {Shoker}, {Shyam}, {Sidor}, {Sigler}, {Simens}, {Sitkin}, {Slama}, {Sohl}, {Sokolowsky}, {Song}, {Staudacher}, {Petroski Such}, {Summers}, {Sutskever}, {Tang}, {Tezak}, {Thompson}, {Tillet}, {Tootoonchian}, {Tseng}, {Tuggle}, {Turley}, {Tworek}, {Cer{\'o}n Uribe}, {Vallone}, {Vijayvergiya}, {Voss}, {Wainwright}, {Wang}, {Wang}, {Wang}, {Ward}, {Wei}, {Weinmann}, {Welihinda}, {Welinder}, {Weng}, {Weng}, {Wiethoff}, {Willner}, {Winter}, {Wolrich}, {Wong}, {Workman}, {Wu}, {Wu}, {Wu}, {Xiao}, {Xu}, {Yoo}, {Yu}, {Yuan}, {Zaremba}, {Zellers}, {Zhang}, {Zhang}, {Zhao}, {Zheng}, {Zhuang}, {Zhuk}, and {Zoph}]{GPT-4}
{OpenAI}, Josh {Achiam}, Steven {Adler}, Sandhini {Agarwal}, Lama {Ahmad}, Ilge {Akkaya}, Florencia {Leoni Aleman}, Diogo {Almeida}, Janko {Altenschmidt}, Sam {Altman}, Shyamal {Anadkat}, Red {Avila}, Igor {Babuschkin}, Suchir {Balaji}, Valerie {Balcom}, Paul {Baltescu}, Haiming {Bao}, Mohammad {Bavarian}, Jeff {Belgum}, Irwan {Bello}, Jake {Berdine}, Gabriel {Bernadett-Shapiro}, Christopher {Berner}, Lenny {Bogdonoff}, Oleg {Boiko}, Madelaine {Boyd}, Anna-Luisa {Brakman}, Greg {Brockman}, Tim {Brooks}, Miles {Brundage}, Kevin {Button}, Trevor {Cai}, Rosie {Campbell}, Andrew {Cann}, Brittany {Carey}, Chelsea {Carlson}, Rory {Carmichael}, Brooke {Chan}, Che {Chang}, Fotis {Chantzis}, Derek {Chen}, Sully {Chen}, Ruby {Chen}, Jason {Chen}, Mark {Chen}, Ben {Chess}, Chester {Cho}, Casey {Chu}, Hyung~Won {Chung}, Dave {Cummings}, Jeremiah {Currier}, Yunxing {Dai}, Cory {Decareaux}, Thomas {Degry}, Noah {Deutsch}, Damien {Deville}, Arka {Dhar}, David {Dohan}, Steve {Dowling}, Sheila {Dunning}, Adrien {Ecoffet},
  Atty {Eleti}, Tyna {Eloundou}, David {Farhi}, Liam {Fedus}, Niko {Felix}, Sim{\'o}n {Posada Fishman}, Juston {Forte}, Isabella {Fulford}, Leo {Gao}, Elie {Georges}, Christian {Gibson}, Vik {Goel}, Tarun {Gogineni}, Gabriel {Goh}, Rapha {Gontijo-Lopes}, Jonathan {Gordon}, Morgan {Grafstein}, Scott {Gray}, Ryan {Greene}, Joshua {Gross}, Shixiang~Shane {Gu}, Yufei {Guo}, Chris {Hallacy}, Jesse {Han}, Jeff {Harris}, Yuchen {He}, Mike {Heaton}, Johannes {Heidecke}, Chris {Hesse}, Alan {Hickey}, Wade {Hickey}, Peter {Hoeschele}, Brandon {Houghton}, Kenny {Hsu}, Shengli {Hu}, Xin {Hu}, Joost {Huizinga}, Shantanu {Jain}, Shawn {Jain}, Joanne {Jang}, Angela {Jiang}, Roger {Jiang}, Haozhun {Jin}, Denny {Jin}, Shino {Jomoto}, Billie {Jonn}, Heewoo {Jun}, Tomer {Kaftan}, {\L}ukasz {Kaiser}, Ali {Kamali}, Ingmar {Kanitscheider}, Nitish {Shirish Keskar}, Tabarak {Khan}, Logan {Kilpatrick}, Jong~Wook {Kim}, Christina {Kim}, Yongjik {Kim}, Jan {Hendrik Kirchner}, Jamie {Kiros}, Matt {Knight}, Daniel {Kokotajlo}, {\L}ukasz
  {Kondraciuk}, Andrew {Kondrich}, Aris {Konstantinidis}, Kyle {Kosic}, Gretchen {Krueger}, Vishal {Kuo}, Michael {Lampe}, Ikai {Lan}, Teddy {Lee}, Jan {Leike}, Jade {Leung}, Daniel {Levy}, Chak~Ming {Li}, Rachel {Lim}, Molly {Lin}, Stephanie {Lin}, Mateusz {Litwin}, Theresa {Lopez}, Ryan {Lowe}, Patricia {Lue}, Anna {Makanju}, Kim {Malfacini}, Sam {Manning}, Todor {Markov}, Yaniv {Markovski}, Bianca {Martin}, Katie {Mayer}, Andrew {Mayne}, Bob {McGrew}, Scott~Mayer {McKinney}, Christine {McLeavey}, Paul {McMillan}, Jake {McNeil}, David {Medina}, Aalok {Mehta}, Jacob {Menick}, Luke {Metz}, Andrey {Mishchenko}, Pamela {Mishkin}, Vinnie {Monaco}, Evan {Morikawa}, Daniel {Mossing}, Tong {Mu}, Mira {Murati}, Oleg {Murk}, David {M{\'e}ly}, Ashvin {Nair}, Reiichiro {Nakano}, Rajeev {Nayak}, Arvind {Neelakantan}, Richard {Ngo}, Hyeonwoo {Noh}, Long {Ouyang}, Cullen {O'Keefe}, Jakub {Pachocki}, Alex {Paino}, Joe {Palermo}, Ashley {Pantuliano}, Giambattista {Parascandolo}, Joel {Parish}, Emy {Parparita}, Alex
  {Passos}, Mikhail {Pavlov}, Andrew {Peng}, Adam {Perelman}, Filipe {de Avila Belbute Peres}, Michael {Petrov}, Henrique {Ponde de Oliveira Pinto}, {Michael}, {Pokorny}, Michelle {Pokrass}, Vitchyr~H. {Pong}, Tolly {Powell}, Alethea {Power}, Boris {Power}, Elizabeth {Proehl}, Raul {Puri}, Alec {Radford}, Jack {Rae}, Aditya {Ramesh}, Cameron {Raymond}, Francis {Real}, Kendra {Rimbach}, Carl {Ross}, Bob {Rotsted}, Henri {Roussez}, Nick {Ryder}, Mario {Saltarelli}, Ted {Sanders}, Shibani {Santurkar}, Girish {Sastry}, Heather {Schmidt}, David {Schnurr}, John {Schulman}, Daniel {Selsam}, Kyla {Sheppard}, Toki {Sherbakov}, Jessica {Shieh}, Sarah {Shoker}, Pranav {Shyam}, Szymon {Sidor}, Eric {Sigler}, Maddie {Simens}, Jordan {Sitkin}, Katarina {Slama}, Ian {Sohl}, Benjamin {Sokolowsky}, Yang {Song}, Natalie {Staudacher}, Felipe {Petroski Such}, Natalie {Summers}, Ilya {Sutskever}, Jie {Tang}, Nikolas {Tezak}, Madeleine~B. {Thompson}, Phil {Tillet}, Amin {Tootoonchian}, Elizabeth {Tseng}, Preston {Tuggle}, Nick
  {Turley}, Jerry {Tworek}, Juan~Felipe {Cer{\'o}n Uribe}, Andrea {Vallone}, Arun {Vijayvergiya}, Chelsea {Voss}, Carroll {Wainwright}, Justin~Jay {Wang}, Alvin {Wang}, Ben {Wang}, Jonathan {Ward}, Jason {Wei}, CJ~{Weinmann}, Akila {Welihinda}, Peter {Welinder}, Jiayi {Weng}, Lilian {Weng}, Matt {Wiethoff}, Dave {Willner}, Clemens {Winter}, Samuel {Wolrich}, Hannah {Wong}, Lauren {Workman}, Sherwin {Wu}, Jeff {Wu}, Michael {Wu}, Kai {Xiao}, Tao {Xu}, Sarah {Yoo}, Kevin {Yu}, Qiming {Yuan}, Wojciech {Zaremba}, Rowan {Zellers}, Chong {Zhang}, Marvin {Zhang}, Shengjia {Zhao}, Tianhao {Zheng}, Juntang {Zhuang}, William {Zhuk}, and Barret {Zoph}.
\newblock {GPT-4 Technical Report}.
\newblock \emph{arXiv e-prints}, art. arXiv:2303.08774, March 2023.
\newblock \doi{10.48550/arXiv.2303.08774}.

\bibitem[Ouyang et~al.(2022)Ouyang, Wu, Jiang, Almeida, Wainwright, Mishkin, Zhang, Agarwal, Slama, Ray, Schulman, Hilton, Kelton, Miller, Simens, Askell, Welinder, Christiano, Leike, and Lowe]{Ouyang_TIHF}
Long Ouyang, Jeffrey Wu, Xu~Jiang, Diogo Almeida, Carroll Wainwright, Pamela Mishkin, Chong Zhang, Sandhini Agarwal, Katarina Slama, Alex Ray, John Schulman, Jacob Hilton, Fraser Kelton, Luke Miller, Maddie Simens, Amanda Askell, Peter Welinder, Paul~F Christiano, Jan Leike, and Ryan Lowe.
\newblock Training language models to follow instructions with human feedback.
\newblock In S.~Koyejo, S.~Mohamed, A.~Agarwal, D.~Belgrave, K.~Cho, and A.~Oh (eds.), \emph{Advances in Neural Information Processing Systems}, volume~35, pp.\  27730--27744. Curran Associates, Inc., 2022.
\newblock URL \url{https://proceedings.neurips.cc/paper_files/paper/2022/file/b1efde53be364a73914f58805a001731-Paper-Conference.pdf}.

\bibitem[Papineni et~al.(2002)Papineni, Roukos, Ward, and Zhu]{BLEU}
Kishore Papineni, Salim Roukos, Todd Ward, and Wei-Jing Zhu.
\newblock {B}leu: a method for automatic evaluation of machine translation.
\newblock In Pierre Isabelle, Eugene Charniak, and Dekang Lin (eds.), \emph{Proceedings of the 40th Annual Meeting of the Association for Computational Linguistics}, pp.\  311--318, Philadelphia, Pennsylvania, USA, July 2002. Association for Computational Linguistics.
\newblock \doi{10.3115/1073083.1073135}.
\newblock URL \url{https://aclanthology.org/P02-1040}.

\bibitem[Raffel et~al.(2020)Raffel, Shazeer, Roberts, Lee, Narang, Matena, Zhou, Li, and Liu]{raffel2020exploring}
Colin Raffel, Noam Shazeer, Adam Roberts, Katherine Lee, Sharan Narang, Michael Matena, Yanqi Zhou, Wei Li, and Peter~J Liu.
\newblock Exploring the limits of transfer learning with a unified text-to-text transformer.
\newblock \emph{Journal of machine learning research}, 21\penalty0 (140):\penalty0 1--67, 2020.

\bibitem[Ram{\'i}rez-S{\'a}nchez et~al.(2020)Ram{\'i}rez-S{\'a}nchez, Zaragoza-Bernabeu, Ba{\~n}{\'o}n, and Rojas]{bicleaner}
Gema Ram{\'i}rez-S{\'a}nchez, Jaume Zaragoza-Bernabeu, Marta Ba{\~n}{\'o}n, and Sergio~Ortiz Rojas.
\newblock Bifixer and bicleaner: two open-source tools to clean your parallel data.
\newblock In Andr{\'e} Martins, Helena Moniz, Sara Fumega, Bruno Martins, Fernando Batista, Luisa Coheur, Carla Parra, Isabel Trancoso, Marco Turchi, Arianna Bisazza, Joss Moorkens, Ana Guerberof, Mary Nurminen, Lena Marg, and Mikel~L. Forcada (eds.), \emph{Proceedings of the 22nd Annual Conference of the European Association for Machine Translation}, pp.\  291--298, Lisboa, Portugal, November 2020. European Association for Machine Translation.
\newblock URL \url{https://aclanthology.org/2020.eamt-1.31/}.

\bibitem[Raunak et~al.(2024)Raunak, Kocmi, and Post]{raunak-etal-2024-slide}
Vikas Raunak, Tom Kocmi, and Matt Post.
\newblock {SLIDE}: Reference-free evaluation for machine translation using a sliding document window.
\newblock In Kevin Duh, Helena Gomez, and Steven Bethard (eds.), \emph{Proceedings of the 2024 Conference of the North American Chapter of the Association for Computational Linguistics: Human Language Technologies (Volume 2: Short Papers)}, pp.\  205--211, Mexico City, Mexico, June 2024. Association for Computational Linguistics.
\newblock \doi{10.18653/v1/2024.naacl-short.18}.
\newblock URL \url{https://aclanthology.org/2024.naacl-short.18}.

\bibitem[Rei et~al.(2022)Rei, C.~de Souza, Alves, Zerva, Farinha, Glushkova, Lavie, Coheur, and Martins]{COMET22}
Ricardo Rei, Jos{\'e}~G. C.~de Souza, Duarte Alves, Chrysoula Zerva, Ana~C Farinha, Taisiya Glushkova, Alon Lavie, Luisa Coheur, and Andr{\'e} F.~T. Martins.
\newblock {COMET}-22: Unbabel-{IST} 2022 submission for the metrics shared task.
\newblock In Philipp Koehn, Lo{\"\i}c Barrault, Ond{\v{r}}ej Bojar, Fethi Bougares, Rajen Chatterjee, Marta~R. Costa-juss{\`a}, Christian Federmann, Mark Fishel, Alexander Fraser, Markus Freitag, Yvette Graham, Roman Grundkiewicz, Paco Guzman, Barry Haddow, Matthias Huck, Antonio Jimeno~Yepes, Tom Kocmi, Andr{\'e} Martins, Makoto Morishita, Christof Monz, Masaaki Nagata, Toshiaki Nakazawa, Matteo Negri, Aur{\'e}lie N{\'e}v{\'e}ol, Mariana Neves, Martin Popel, Marco Turchi, and Marcos Zampieri (eds.), \emph{Proceedings of the Seventh Conference on Machine Translation (WMT)}, pp.\  578--585, Abu Dhabi, United Arab Emirates (Hybrid), December 2022. Association for Computational Linguistics.
\newblock URL \url{https://aclanthology.org/2022.wmt-1.52}.

\bibitem[{Rei} et~al.(2023){Rei}, {Guerreiro}, {Pombal}, {van Stigt}, {Treviso}, {Coheur}, {de Souza}, and {Martins}]{cometkiwi23}
Ricardo {Rei}, Nuno~M. {Guerreiro}, Jos{\'e} {Pombal}, Daan {van Stigt}, Marcos {Treviso}, Luisa {Coheur}, Jos{\'e} G.~C. {de Souza}, and Andr{\'e} F.~T. {Martins}.
\newblock {Scaling up COMETKIWI: Unbabel-IST 2023 Submission for the Quality Estimation Shared Task}.
\newblock \emph{arXiv e-prints}, art. arXiv:2309.11925, September 2023.
\newblock \doi{10.48550/arXiv.2309.11925}.

\bibitem[{Sanh} et~al.(2021){Sanh}, {Webson}, {Raffel}, {Bach}, {Sutawika}, {Alyafeai}, {Chaffin}, {Stiegler}, {Le Scao}, {Raja}, {Dey}, {Saiful Bari}, {Xu}, {Thakker}, {Sharma Sharma}, {Szczechla}, {Kim}, {Chhablani}, {Nayak}, {Datta}, {Chang}, {Tian-Jian Jiang}, {Wang}, {Manica}, {Shen}, {Yong}, {Pandey}, {Bawden}, {Wang}, {Neeraj}, {Rozen}, {Sharma}, {Santilli}, {Fevry}, {Fries}, {Teehan}, {Bers}, {Biderman}, {Gao}, {Wolf}, and {Rush}]{Sanh_0STG}
Victor {Sanh}, Albert {Webson}, Colin {Raffel}, Stephen~H. {Bach}, Lintang {Sutawika}, Zaid {Alyafeai}, Antoine {Chaffin}, Arnaud {Stiegler}, Teven {Le Scao}, Arun {Raja}, Manan {Dey}, M~{Saiful Bari}, Canwen {Xu}, Urmish {Thakker}, Shanya {Sharma Sharma}, Eliza {Szczechla}, Taewoon {Kim}, Gunjan {Chhablani}, Nihal {Nayak}, Debajyoti {Datta}, Jonathan {Chang}, Mike {Tian-Jian Jiang}, Han {Wang}, Matteo {Manica}, Sheng {Shen}, Zheng~Xin {Yong}, Harshit {Pandey}, Rachel {Bawden}, Thomas {Wang}, Trishala {Neeraj}, Jos {Rozen}, Abheesht {Sharma}, Andrea {Santilli}, Thibault {Fevry}, Jason~Alan {Fries}, Ryan {Teehan}, Tali {Bers}, Stella {Biderman}, Leo {Gao}, Thomas {Wolf}, and Alexander~M. {Rush}.
\newblock {Multitask Prompted Training Enables Zero-Shot Task Generalization}.
\newblock \emph{arXiv e-prints}, art. arXiv:2110.08207, October 2021.
\newblock \doi{10.48550/arXiv.2110.08207}.

\bibitem[{Su} et~al.(2021){Su}, {Lu}, {Pan}, {Murtadha}, {Wen}, and {Liu}]{ROPE}
Jianlin {Su}, Yu~{Lu}, Shengfeng {Pan}, Ahmed {Murtadha}, Bo~{Wen}, and Yunfeng {Liu}.
\newblock {RoFormer: Enhanced Transformer with Rotary Position Embedding}.
\newblock \emph{arXiv e-prints}, art. arXiv:2104.09864, April 2021.
\newblock \doi{10.48550/arXiv.2104.09864}.

\bibitem[Sun et~al.(2022)Sun, Wang, Zhou, Zhao, Huang, Chen, and Li]{sun-etal-2022-rethinking}
Zewei Sun, Mingxuan Wang, Hao Zhou, Chengqi Zhao, Shujian Huang, Jiajun Chen, and Lei Li.
\newblock Rethinking document-level neural machine translation.
\newblock In Smaranda Muresan, Preslav Nakov, and Aline Villavicencio (eds.), \emph{Findings of the Association for Computational Linguistics: ACL 2022}, pp.\  3537--3548, Dublin, Ireland, May 2022. Association for Computational Linguistics.
\newblock \doi{10.18653/v1/2022.findings-acl.279}.
\newblock URL \url{https://aclanthology.org/2022.findings-acl.279}.

\bibitem[Sutskever et~al.(2014)Sutskever, Vinyals, and Le]{Sutskever_2014}
Ilya Sutskever, Oriol Vinyals, and Quoc~V Le.
\newblock Sequence to sequence learning with neural networks.
\newblock In Z.~Ghahramani, M.~Welling, C.~Cortes, N.~Lawrence, and K.Q. Weinberger (eds.), \emph{Advances in Neural Information Processing Systems}, volume~27. Curran Associates, Inc., 2014.
\newblock URL \url{https://proceedings.neurips.cc/paper_files/paper/2014/file/a14ac55a4f27472c5d894ec1c3c743d2-Paper.pdf}.

\bibitem[Toral(2020)]{TORAL}
Antonio Toral.
\newblock Reassessing claims of human parity and super-human performance in machine translation at {WMT} 2019.
\newblock In Andr{\'e} Martins, Helena Moniz, Sara Fumega, Bruno Martins, Fernando Batista, Luisa Coheur, Carla Parra, Isabel Trancoso, Marco Turchi, Arianna Bisazza, Joss Moorkens, Ana Guerberof, Mary Nurminen, Lena Marg, and Mikel~L. Forcada (eds.), \emph{Proceedings of the 22nd Annual Conference of the European Association for Machine Translation}, pp.\  185--194, Lisboa, Portugal, November 2020. European Association for Machine Translation.
\newblock URL \url{https://aclanthology.org/2020.eamt-1.20}.

\bibitem[{Toral} \& {Way}(2018){Toral} and {Way}]{LITERARY-2}
Antonio {Toral} and Andy {Way}.
\newblock {What Level of Quality can Neural Machine Translation Attain on Literary Text?}
\newblock \emph{arXiv e-prints}, art. arXiv:1801.04962, January 2018.
\newblock \doi{10.48550/arXiv.1801.04962}.

\bibitem[{Touvron} et~al.(2023){Touvron}, {Lavril}, {Izacard}, {Martinet}, {Lachaux}, {Lacroix}, {Rozi{\`e}re}, {Goyal}, {Hambro}, {Azhar}, {Rodriguez}, {Joulin}, {Grave}, and {Lample}]{LLAMA}
Hugo {Touvron}, Thibaut {Lavril}, Gautier {Izacard}, Xavier {Martinet}, Marie-Anne {Lachaux}, Timoth{\'e}e {Lacroix}, Baptiste {Rozi{\`e}re}, Naman {Goyal}, Eric {Hambro}, Faisal {Azhar}, Aurelien {Rodriguez}, Armand {Joulin}, Edouard {Grave}, and Guillaume {Lample}.
\newblock {LLaMA: Open and Efficient Foundation Language Models}.
\newblock \emph{arXiv e-prints}, art. arXiv:2302.13971, February 2023.
\newblock \doi{10.48550/arXiv.2302.13971}.

\bibitem[Vaswani et~al.(2017)Vaswani, Shazeer, Parmar, Uszkoreit, Jones, Gomez, Kaiser, and Polosukhin]{Vaswani_2017}
Ashish Vaswani, Noam Shazeer, Niki Parmar, Jakob Uszkoreit, Llion Jones, Aidan~N Gomez, \L~ukasz Kaiser, and Illia Polosukhin.
\newblock Attention is all you need.
\newblock In I.~Guyon, U.~Von Luxburg, S.~Bengio, H.~Wallach, R.~Fergus, S.~Vishwanathan, and R.~Garnett (eds.), \emph{Advances in Neural Information Processing Systems}, volume~30. Curran Associates, Inc., 2017.
\newblock URL \url{https://proceedings.neurips.cc/paper_files/paper/2017/file/3f5ee243547dee91fbd053c1c4a845aa-Paper.pdf}.

\bibitem[Vernikos et~al.(2022)Vernikos, Thompson, Mathur, and Federico]{vernikos-etal-2022-embarrassingly}
Giorgos Vernikos, Brian Thompson, Prashant Mathur, and Marcello Federico.
\newblock Embarrassingly easy document-level {MT} metrics: How to convert any pretrained metric into a document-level metric.
\newblock In Philipp Koehn, Lo{\"\i}c Barrault, Ond{\v{r}}ej Bojar, Fethi Bougares, Rajen Chatterjee, Marta~R. Costa-juss{\`a}, Christian Federmann, Mark Fishel, Alexander Fraser, Markus Freitag, Yvette Graham, Roman Grundkiewicz, Paco Guzman, Barry Haddow, Matthias Huck, Antonio Jimeno~Yepes, Tom Kocmi, Andr{\'e} Martins, Makoto Morishita, Christof Monz, Masaaki Nagata, Toshiaki Nakazawa, Matteo Negri, Aur{\'e}lie N{\'e}v{\'e}ol, Mariana Neves, Martin Popel, Marco Turchi, and Marcos Zampieri (eds.), \emph{Proceedings of the Seventh Conference on Machine Translation (WMT)}, pp.\  118--128, Abu Dhabi, United Arab Emirates (Hybrid), December 2022. Association for Computational Linguistics.
\newblock URL \url{https://aclanthology.org/2022.wmt-1.6}.

\bibitem[Voita et~al.(2018)Voita, Serdyukov, Sennrich, and Titov]{ME-2}
Elena Voita, Pavel Serdyukov, Rico Sennrich, and Ivan Titov.
\newblock Context-aware neural machine translation learns anaphora resolution.
\newblock In Iryna Gurevych and Yusuke Miyao (eds.), \emph{Proceedings of the 56th Annual Meeting of the Association for Computational Linguistics (Volume 1: Long Papers)}, pp.\  1264--1274, Melbourne, Australia, July 2018. Association for Computational Linguistics.
\newblock \doi{10.18653/v1/P18-1117}.
\newblock URL \url{https://aclanthology.org/P18-1117}.

\bibitem[Wang et~al.(2023{\natexlab{a}})Wang, Lyu, Ji, Zhang, Yu, Shi, and Tu]{DocMT_WANG}
Longyue Wang, Chenyang Lyu, Tianbo Ji, Zhirui Zhang, Dian Yu, Shuming Shi, and Zhaopeng Tu.
\newblock Document-level machine translation with large language models.
\newblock In Houda Bouamor, Juan Pino, and Kalika Bali (eds.), \emph{Proceedings of the 2023 Conference on Empirical Methods in Natural Language Processing}, pp.\  16646--16661, Singapore, December 2023{\natexlab{a}}. Association for Computational Linguistics.
\newblock \doi{10.18653/v1/2023.emnlp-main.1036}.
\newblock URL \url{https://aclanthology.org/2023.emnlp-main.1036}.

\bibitem[Wang et~al.(2023{\natexlab{b}})Wang, Tu, Gu, Liu, Yu, Ma, Lyu, Zhou, Liu, Ma, Chen, Graham, Webber, Koehn, Way, Yuan, and Shi]{GuoFeng_ST}
Longyue Wang, Zhaopeng Tu, Yan Gu, Siyou Liu, Dian Yu, Qingsong Ma, Chenyang Lyu, Liting Zhou, Chao-Hong Liu, Yufeng Ma, Weiyu Chen, Yvette Graham, Bonnie Webber, Philipp Koehn, Andy Way, Yulin Yuan, and Shuming Shi.
\newblock Findings of the {WMT} 2023 shared task on discourse-level literary translation: A fresh orb in the cosmos of {LLM}s.
\newblock In Philipp Koehn, Barry Haddow, Tom Kocmi, and Christof Monz (eds.), \emph{Proceedings of the Eighth Conference on Machine Translation}, pp.\  55--67, Singapore, December 2023{\natexlab{b}}. Association for Computational Linguistics.
\newblock \doi{10.18653/v1/2023.wmt-1.3}.
\newblock URL \url{https://aclanthology.org/2023.wmt-1.3}.

\bibitem[{Wang} et~al.(2024){Wang}, {Zeng}, {Liu}, {Wong}, {Meng}, {Zhou}, and {Zhang}]{DELTA_LLM}
Yutong {Wang}, Jiali {Zeng}, Xuebo {Liu}, Derek~F. {Wong}, Fandong {Meng}, Jie {Zhou}, and Min {Zhang}.
\newblock {DelTA: An Online Document-Level Translation Agent Based on Multi-Level Memory}.
\newblock \emph{arXiv e-prints}, art. arXiv:2410.08143, October 2024.
\newblock \doi{10.48550/arXiv.2410.08143}.

\bibitem[{Wei} et~al.(2021){Wei}, {Bosma}, {Zhao}, {Guu}, {Yu}, {Lester}, {Du}, {Dai}, and {Le}]{Wei_0SL}
Jason {Wei}, Maarten {Bosma}, Vincent~Y. {Zhao}, Kelvin {Guu}, Adams~Wei {Yu}, Brian {Lester}, Nan {Du}, Andrew~M. {Dai}, and Quoc~V. {Le}.
\newblock {Finetuned Language Models Are Zero-Shot Learners}.
\newblock \emph{arXiv e-prints}, art. arXiv:2109.01652, September 2021.
\newblock \doi{10.48550/arXiv.2109.01652}.

\bibitem[Wong \& Kit(2012)Wong and Kit]{DocMetric-2}
Billy T.~M. Wong and Chunyu Kit.
\newblock Extending machine translation evaluation metrics with lexical cohesion to document level.
\newblock In Jun{'}ichi Tsujii, James Henderson, and Marius Pa{\c{s}}ca (eds.), \emph{Proceedings of the 2012 Joint Conference on Empirical Methods in Natural Language Processing and Computational Natural Language Learning}, pp.\  1060--1068, Jeju Island, Korea, July 2012. Association for Computational Linguistics.
\newblock URL \url{https://aclanthology.org/D12-1097}.

\bibitem[Wu et~al.(2023)Wu, Foster, Qu, and Haffari]{ATTENTION-3}
Minghao Wu, George Foster, Lizhen Qu, and Gholamreza Haffari.
\newblock Document flattening: Beyond concatenating context for document-level neural machine translation.
\newblock In Andreas Vlachos and Isabelle Augenstein (eds.), \emph{Proceedings of the 17th Conference of the European Chapter of the Association for Computational Linguistics}, pp.\  448--462, Dubrovnik, Croatia, May 2023. Association for Computational Linguistics.
\newblock \doi{10.18653/v1/2023.eacl-main.33}.
\newblock URL \url{https://aclanthology.org/2023.eacl-main.33}.

\bibitem[{Wu} et~al.(2024){Wu}, {Vu}, {Qu}, {Foster}, and {Haffari}]{DocMT_WU}
Minghao {Wu}, Thuy-Trang {Vu}, Lizhen {Qu}, George {Foster}, and Gholamreza {Haffari}.
\newblock {Adapting Large Language Models for Document-Level Machine Translation}.
\newblock \emph{arXiv e-prints}, art. arXiv:2401.06468, January 2024.
\newblock \doi{10.48550/arXiv.2401.06468}.

\bibitem[Wu et~al.(2024{\natexlab{a}})Wu, Wang, Foster, Qu, and Haffari]{wu-etal-2024-importance}
Minghao Wu, Yufei Wang, George Foster, Lizhen Qu, and Gholamreza Haffari.
\newblock Importance-aware data augmentation for document-level neural machine translation.
\newblock In Yvette Graham and Matthew Purver (eds.), \emph{Proceedings of the 18th Conference of the European Chapter of the Association for Computational Linguistics (Volume 1: Long Papers)}, pp.\  740--752, St. Julian{'}s, Malta, March 2024{\natexlab{a}}. Association for Computational Linguistics.
\newblock \doi{10.18653/v1/2024.eacl-long.44}.
\newblock URL \url{https://aclanthology.org/2024.eacl-long.44/}.

\bibitem[Wu et~al.(2024{\natexlab{b}})Wu, Yuan, Haffari, and Wang]{AGENT_LLM}
Minghao Wu, Yulin Yuan, Gholamreza Haffari, and Longyue Wang.
\newblock (perhaps) beyond human translation: Harnessing multi-agent collaboration for translating ultra-long literary texts, 2024{\natexlab{b}}.
\newblock URL \url{https://arxiv.org/abs/2405.11804}.

\bibitem[Xiao et~al.(2011)Xiao, Zhu, Yao, and Zhang]{xiao-etal-2011-document}
Tong Xiao, Jingbo Zhu, Shujie Yao, and Hao Zhang.
\newblock Document-level consistency verification in machine translation.
\newblock In \emph{Proceedings of Machine Translation Summit XIII: Papers}, Xiamen, China, September 19-23 2011.
\newblock URL \url{https://aclanthology.org/2011.mtsummit-papers.13}.

\bibitem[Xie et~al.(2023)Xie, Li, Wu, Wei, Chen, Rao, Li, Shang, Guo, Lei, Yang, and Jiang]{xie-etal-2023-hw}
Yuhao Xie, Zongyao Li, Zhanglin Wu, Daimeng Wei, Xiaoyu Chen, Zhiqiang Rao, Shaojun Li, Hengchao Shang, Jiaxin Guo, Lizhi Lei, Hao Yang, and Yanfei Jiang.
\newblock {HW}-{TSC}`s submissions to the {WMT}23 discourse-level literary translation shared task.
\newblock In Philipp Koehn, Barry Haddow, Tom Kocmi, and Christof Monz (eds.), \emph{Proceedings of the Eighth Conference on Machine Translation}, pp.\  302--306, Singapore, December 2023. Association for Computational Linguistics.
\newblock \doi{10.18653/v1/2023.wmt-1.32}.
\newblock URL \url{https://aclanthology.org/2023.wmt-1.32/}.

\bibitem[{Xiong} et~al.(2018){Xiong}, {He}, {Wu}, and {Wang}]{DISCOURSE_STRUCT_1}
Hao {Xiong}, Zhongjun {He}, Hua {Wu}, and Haifeng {Wang}.
\newblock {Modeling Coherence for Discourse Neural Machine Translation}.
\newblock \emph{arXiv e-prints}, art. arXiv:1811.05683, November 2018.
\newblock \doi{10.48550/arXiv.1811.05683}.

\bibitem[{Xu} et~al.(2023){Xu}, {Kim}, {Sharaf}, and {Awadalla}]{XU_PARADIGM}
Haoran {Xu}, Young~Jin {Kim}, Amr {Sharaf}, and Hany~Hassan {Awadalla}.
\newblock {A Paradigm Shift in Machine Translation: Boosting Translation Performance of Large Language Models}.
\newblock \emph{arXiv e-prints}, art. arXiv:2309.11674, September 2023.
\newblock \doi{10.48550/arXiv.2309.11674}.

\bibitem[Zhang et~al.(2023)Zhang, Haddow, and Birch]{LLM_PROMPT_MT}
Biao Zhang, Barry Haddow, and Alexandra Birch.
\newblock Prompting large language model for machine translation: A case study.
\newblock In Andreas Krause, Emma Brunskill, Kyunghyun Cho, Barbara Engelhardt, Sivan Sabato, and Jonathan Scarlett (eds.), \emph{Proceedings of the 40th International Conference on Machine Learning}, volume 202 of \emph{Proceedings of Machine Learning Research}, pp.\  41092--41110. PMLR, 23--29 Jul 2023.
\newblock URL \url{https://proceedings.mlr.press/v202/zhang23m.html}.

\bibitem[Zhang et~al.(2018)Zhang, Luan, Sun, Zhai, Xu, Zhang, and Liu]{ME-1}
Jiacheng Zhang, Huanbo Luan, Maosong Sun, Feifei Zhai, Jingfang Xu, Min Zhang, and Yang Liu.
\newblock Improving the transformer translation model with document-level context.
\newblock In Ellen Riloff, David Chiang, Julia Hockenmaier, and Jun{'}ichi Tsujii (eds.), \emph{Proceedings of the 2018 Conference on Empirical Methods in Natural Language Processing}, pp.\  533--542, Brussels, Belgium, October-November 2018. Association for Computational Linguistics.
\newblock \doi{10.18653/v1/D18-1049}.
\newblock URL \url{https://aclanthology.org/D18-1049}.

\bibitem[Zhang et~al.(2020)Zhang, Chen, Ge, and Fan]{ATTENTION-1}
Pei Zhang, Boxing Chen, Niyu Ge, and Kai Fan.
\newblock Long-short term masking transformer: A simple but effective baseline for document-level neural machine translation.
\newblock In Bonnie Webber, Trevor Cohn, Yulan He, and Yang Liu (eds.), \emph{Proceedings of the 2020 Conference on Empirical Methods in Natural Language Processing (EMNLP)}, pp.\  1081--1087, Online, November 2020. Association for Computational Linguistics.
\newblock \doi{10.18653/v1/2020.emnlp-main.81}.
\newblock URL \url{https://aclanthology.org/2020.emnlp-main.81}.

\bibitem[Ziemski et~al.(2016)Ziemski, Junczys-Dowmunt, and Pouliquen]{UNPC}
Micha{\l} Ziemski, Marcin Junczys-Dowmunt, and Bruno Pouliquen.
\newblock The {U}nited {N}ations parallel corpus v1.0.
\newblock In Nicoletta Calzolari, Khalid Choukri, Thierry Declerck, Sara Goggi, Marko Grobelnik, Bente Maegaard, Joseph Mariani, Helene Mazo, Asuncion Moreno, Jan Odijk, and Stelios Piperidis (eds.), \emph{Proceedings of the Tenth International Conference on Language Resources and Evaluation ({LREC}'16)}, pp.\  3530--3534, Portoro{\v{z}}, Slovenia, May 2016. European Language Resources Association (ELRA).
\newblock URL \url{https://aclanthology.org/L16-1561}.

\end{thebibliography}

\clearpage
\appendix

\section{Document-level Training Hyperparameters}
\label{sec:docmt_hyperparams}

Table~\ref{tab:docmt_hyperparams} contains the hyperparameter configuration for the training of DocMT-LLMs.

\begin{table}[!h]
\renewcommand{\arraystretch}{1.1}
\begin{center}
\begin{tabular}{ll}
\toprule
Batch size & $32$ \\
Number of Epochs & 2 \\
Learning rate & $7 \times 10^{-6}$ \\
LR Scheduler & cosine \\
Warmup Steps & $125$ \\
Weight Decay & $0.01 $\\
Optimizer & Adam \citep{2014arXiv1412.6980K} \\
Adam $\beta_1$ & $0.9$ \\
Adam $\beta_2$ & $0.999$ \\
Adam $\epsilon$ & $1 \times 10^{-8}$ \\
Maximum Sequence Length & $32768$ \\
\bottomrule
\end{tabular}    
\end{center}
\caption{Hyperparameter configuration to fine-tune DocMT-LLMs on \docblocks.}
\label{tab:docmt_hyperparams}
\end{table}

\section{Full Translation Results}
\label{sec:full_mt_results}

In this appendix, we provide detailed sentence-level and document-level evaluation results, broken down by individual language pairs.
Tables~\ref{tab:iwslt-dbleu}, \ref{tab:iwslt-dcomet}, and \ref{tab:iwslt-ltcr} report document-level evaluation results across IWSLT2017 language pairs using d-\textsc{BLEU}, d-\textsc{COMET}, and \textsc{LTCR}, respectively \---\ similar to the results presented in Table~\ref{tab:iwslt_doc2doc}.
Additionally, Tables~\ref{tab:wmt23-comet-app}, \ref{tab:tico19-comet-app}, and \ref{tab:flores-comet-app} present the same comparisons as Table~\ref{tab:SentMT} in the main paper, illustrating the impact of document-level training on sentence-level translation performance across diverse datasets and language directions.

\begin{table}[h]
\centering
\footnotesize
\scalebox{0.81}{
\renewcommand{\arraystretch}{1.3} 
\begin{tabular}{lccccccccc}
\toprule
\multirow{2}{*}{\textbf{Models}} & \multicolumn{9}{c}{\textbf{IWSLT2017 (en$\rightarrow$xx)}} \\
 & \textbf{de} & \textbf{es} & \textbf{fr} & \textbf{it} & \textbf{ko} & \textbf{nl} & \textbf{pt} & \textbf{ru} & \textbf{zh} \\
\midrule
\texttt{Llama-3.3-70B-Instruct} & \underline{30.51} & 35.17 & 34.45 & 28.96 & 19.59 & \textbf{\underline{32.20}} & 25.08 & 15.12 & 22.64 \\
\texttt{Qwen2.5-72B-Instruct} & 24.86 & 34.34 & \underline{34.63} & 28.65 & 20.54 & 30.73 & 24.74 & 13.50 & \underline{27.32} \\
\texttt{GPT-4o} & 30.49 & \textbf{\underline{50.21}} & 33.51 & \textbf{\underline{30.29}} & \textbf{\underline{22.71}} & 30.39 & \textbf{\underline{42.21}} & \textbf{\underline{35.21}} & 15.21 \\
\midrule
\texttt{TowerInstruct-Mistral-7B} & 2.63 & 2.07 & 3.82 & 2.36 & 0.41 & 3.00 & 4.12 & 1.83 & 1.44 \\
\texttt{DocMT-TowerInstruct-Mistral-7B} & \underline{28.58} & \underline{23.92} & \textbf{\underline{35.45}} & \underline{29.34} & \underline{18.59} & \underline{26.51} & \underline{27.72} & \underline{13.15} & \underline{27.13} \\
\midrule
\texttt{EuroLLM-9B-Instruct} & 4.42 & 5.53 & 6.65 & 3.45 & 0.33 & 3.29 & 3.75 & 4.69 & 1.64 \\
\texttt{DocMT-EuroLLM-9B-Instruct} & \underline{21.21} & \underline{32.19} & \underline{19.91} & \underline{15.98} & \underline{14.30} & \underline{18.98} & \underline{29.62} & \underline{12.49} & \underline{17.48} \\
\midrule
\texttt{Qwen2.5-7B-Instruct} & 26.51 & 31.88 & 30.20 & \underline{24.80} & 0.27 & \underline{29.28} & \underline{23.35} & \underline{9.84} & 11.62 \\
\texttt{DocMT-Qwen2.5-7B-Instruct} & \textbf{\underline{31.08}} & \underline{33.59} & \underline{34.85} & 23.19 & \underline{19.43} & 27.83 & 20.52 & 8.09 & \textbf{\underline{29.03}} \\
\toprule
\multirow{2}{*}{\textbf{Models}} & \multicolumn{9}{c}{\textbf{IWSLT (xx$\rightarrow$en)}} \\
 & \textbf{de} & \textbf{es} & \textbf{fr} & \textbf{it} & \textbf{ko} & \textbf{nl} & \textbf{pt} & \textbf{ru} & \textbf{zh} \\
\midrule
\texttt{Llama-3.3-70B-Instruct} & 46.02 & \textbf{\underline{45.55}} & 45.56 & \textbf{\underline{45.21}} & 16.63 & \textbf{\underline{48.87}} & \textbf{\underline{40.60}} & \underline{28.75} & 24.01 \\
\texttt{Qwen2.5-72B-Instruct} & \textbf{\underline{46.07}} & 44.11 & \underline{45.66} & 44.86 & \underline{24.86} & 48.52 & 40.06 & 28.20 & \underline{26.96} \\
\texttt{GPT-4o} & 42.36 & 40.36 & 41.36 & 39.36 & 13.36 & 43.36 & 36.36 & 23.36 & 21.36 \\
\midrule
\texttt{TowerInstruct-Mistral-7B} & 3.15 & 2.49 & 3.57 & 2.94 & 0.78 & 4.08 & 2.97 & 0.95 & 0.32 \\
\texttt{DocMT-TowerInstruct-Mistral-7B} & \underline{18.38} & \underline{31.98} & \underline{23.01} & \underline{38.61} & \underline{30.79} & \underline{19.31} & \underline{27.99} & \underline{17.84} & \textbf{\underline{33.20}} \\
\midrule
\texttt{EuroLLM-9B-Instruct} & 6.31 & 0.55 & 6.86 & 4.29 & 3.74 & 4.46 & 1.91 & 4.66 & 2.95 \\
\texttt{DocMT-EuroLLM-9B-Instruct} & \underline{31.92} & \underline{33.87} & \underline{21.32} & \underline{22.43} & \underline{28.35} & \underline{37.94} & \underline{30.41} & \underline{18.37} & \underline{31.54} \\
\midrule
\texttt{Qwen2.5-7B-Instruct} & 40.78 & 40.28 & 41.39 & 41.38 & 20.83 & 41.15 & 36.61 & 24.70 & 24.18 \\
\texttt{DocMT-Qwen2.5-7B-Instruct} & \underline{46.00} & \underline{44.05} & \textbf{\underline{47.33}} & \underline{44.93} & \textbf{\underline{36.94}} & \underline{46.45} & \underline{37.88} & \textbf{\underline{30.35}} & \underline{31.11} \\
\bottomrule
\end{tabular}
}
\caption{Evaluation based on d-\textsc{BLEU} across IWSLT2017 language pairs.}
\label{tab:iwslt-dbleu}
\end{table}

\begin{table}[h]
\centering
\scalebox{0.81}{
\renewcommand{\arraystretch}{1.3} 
\begin{tabular}{lccccccccc}
\toprule
\multirow{2}{*}{\textbf{Models}} & \multicolumn{9}{c}{\textbf{IWSLT2017 (en$\rightarrow$xx)}} \\
 & \textbf{de} & \textbf{es} & \textbf{fr} & \textbf{it} & \textbf{ko} & \textbf{nl} & \textbf{pt} & \textbf{ru} & \textbf{zh} \\
\midrule
\texttt{Llama-3.3-70B-Instruct} & 67.90 & 71.31 & 66.91 & 69.81 & 61.11 & 72.49 & 70.33 & 74.00 & 81.18 \\
\texttt{Qwen2.5-72B-Instruct} & 71.26 & \textbf{\underline{73.16}} & \underline{69.50} & 72.49 & \textbf{\underline{81.45}} & 74.55 & \underline{72.03} & \textbf{\underline{77.82}} & 84.08 \\
\texttt{GPT-4o} & \underline{74.90} & 72.38 & 64.67 & \underline{75.39} & 80.69 & \underline{76.49} & 69.38 & 74.38 & \textbf{\underline{84.30}} \\
\midrule
\texttt{TowerInstruct-Mistral-7B} & 27.78 & 28.63 & 30.04 & 27.18 & 30.57 & 27.29 & 30.41 & 30.72 & 34.50 \\
\texttt{DocMT-TowerInstruct-Mistral-7B} & \textbf{\underline{76.39}} & \underline{63.74} & \textbf{\underline{72.47}} & \textbf{\underline{77.51}} & \underline{74.62} & \textbf{\underline{79.13}} & \textbf{\underline{74.04}} & \underline{75.88} & \underline{75.64} \\
\midrule
\texttt{EuroLLM-9B-Instruct} & 20.70 & 30.50 & 20.49 & 24.24 & 20.72 & 23.78 & 29.04 & 33.26 & 21.99 \\
\texttt{DocMT-EuroLLM-9B-Instruct} & \underline{64.01} & \underline{66.37} & \underline{58.64} & \underline{63.91} & \underline{57.12} & \underline{67.94} & \underline{68.34} & \underline{69.65} & \underline{39.68} \\
\midrule
\texttt{Qwen2.5-7B-Instruct} & 68.92 & 71.42 & 66.51 & 68.39 & 76.40 & 69.91 & \underline{69.49} & 71.45 & 83.08 \\
\texttt{DocMT-Qwen2.5-7B-Instruct} & \underline{71.13} & \underline{72.75} & \underline{68.55} & \underline{72.59} & \underline{77.39} & \underline{75.15} & 64.40 & \underline{71.84} & \underline{83.21} \\
\toprule
\multirow{2}{*}{\textbf{Models}} & \multicolumn{9}{c}{\textbf{IWSLT2017 (xx$\rightarrow$en)}} \\
 & \textbf{de} & \textbf{es} & \textbf{fr} & \textbf{it} & \textbf{ko} & \textbf{nl} & \textbf{pt} & \textbf{ru} & \textbf{zh} \\
\midrule
\texttt{Llama-3.3-70B-Instruct} & 71.14 & 72.82 & \underline{72.24} & 69.97 & 78.11 & 67.30 & 69.29 & 73.97 & 73.24 \\
\texttt{Qwen2.5-72B-Instruct} & 68.95 & 70.49 & 65.72 & 62.37 & 70.45 & 64.92 & \underline{71.12} & 70.39 & 71.55 \\
\texttt{GPT-4o} & \textbf{\underline{75.74}} & \textbf{\underline{73.45}} & 66.79 & \textbf{\underline{76.45}} & \underline{78.95} & \textbf{\underline{77.25}} & 67.89 & \textbf{\underline{75.60}} & \textbf{\underline{83.25}} \\
\midrule
\texttt{TowerInstruct-Mistral-7B} & 34.08 & 32.01 & 30.77 & 33.88 & 30.93 & 33.34 & 31.87 & 24.63 & 34.59 \\
\texttt{DocMT-TowerInstruct-Mistral-7B} & \underline{55.78} & \underline{68.36} & \underline{51.34} & \underline{64.15} & \underline{54.80} & \underline{54.30} & \underline{69.13} & \underline{68.47} & \underline{56.30} \\
\midrule
\texttt{EuroLLM-9B-Instruct} & 29.38 & 30.10 & 28.25 & 29.48 & 29.29 & 27.55 & 32.17 & 30.90 & 31.15 \\
\texttt{DocMT-EuroLLM-9B-Instruct} & \underline{63.62} & \underline{70.64} & \underline{53.36} & \underline{68.65} & \underline{44.61} & \underline{67.55} & \underline{72.65} & \underline{71.65} & \underline{51.67} \\
\midrule
\texttt{Qwen2.5-7B-Instruct} & 72.54 & \underline{71.90} & 71.66 & 73.81 & 77.42 & 74.23 & 73.40 & \underline{73.47} & \underline{71.17} \\
\texttt{DocMT-Qwen2.5-7B-Instruct} & \underline{74.63} & 63.36 & \textbf{\underline{73.28}} & \underline{74.80} & \textbf{\underline{79.14}} & \underline{75.39} & \textbf{\underline{74.33}} & 72.64 & 69.60 \\
\bottomrule
\end{tabular}
}
\caption{Evaluation based on d-\textsc{COMET} across IWSLT2017 language pairs.}
\label{tab:iwslt-dcomet}
\end{table}

\begin{table}[h]
\centering
\scalebox{0.81}{
\renewcommand{\arraystretch}{1.3} 
\begin{tabular}{lccccccccc}
\toprule
\multirow{2}{*}{\textbf{Models}} & \multicolumn{9}{c}{\textbf{IWSLT2017 (en$\rightarrow$xx)}} \\
 & \textbf{de} & \textbf{es} & \textbf{fr} & \textbf{it} & \textbf{ko} & \textbf{nl} & \textbf{pt} & \textbf{ru} & \textbf{zh} \\
\midrule
\texttt{Llama-3.3-70B-Instruct} & 60.37 & 76.33 & 61.01 & 64.43 & 24.09 & 58.58 & 62.65 & \textbf{\underline{56.77}} & 54.52 \\
\texttt{Qwen2.5-72B-Instruct} & 62.41 & \textbf{\underline{79.40}} & \underline{61.25} & \underline{66.03} & 41.19 & 59.35 & \underline{65.65} & 56.23 & \textbf{\underline{55.24}} \\
\texttt{GPT-4o} & \textbf{\underline{71.83}} & 61.97 & 60.32 & 65.66 & \textbf{\underline{56.01}} & \underline{61.75} & 57.97 & 48.97 & 46.97 \\
\midrule
\texttt{TowerInstruct-Mistral-7B} & 25.96 & 26.55 & 24.68 & 44.73 & 26.17 & 27.10 & 29.06 & 19.27 & 24.24 \\
\texttt{DocMT-TowerInstruct-Mistral-7B} & \underline{59.93} & \underline{72.67} & \textbf{\underline{64.31}} & \underline{63.59} & \underline{41.62} & \underline{66.95} & \textbf{\underline{68.64}} & \underline{55.77} & \underline{50.66} \\
\midrule
\texttt{EuroLLM-9B-Instruct} & 45.05 & 33.63 & 48.45 & 48.25 & 26.78 & 37.82 & 26.75 & 17.58 & 32.48 \\
\texttt{DocMT-EuroLLM-9B-Instruct} & \underline{56.79} & \underline{73.23} & \underline{60.23} & \underline{62.60} & \underline{31.65} & \underline{47.98} & \underline{68.51} & \underline{48.13} & \underline{50.11} \\
\midrule
\texttt{Qwen2.5-7B-Instruct} & 57.74 & 70.77 & 59.10 & 61.28 & 37.16 & 54.79 & 65.87 & 47.87 & 52.66 \\
\texttt{DocMT-Qwen2.5-7B-Instruct} & \underline{61.21} & \underline{76.78} & \underline{62.86} & \textbf{\underline{71.58}} & \underline{45.22} & \textbf{\underline{67.71}} & \underline{68.25} & \underline{50.32} & \underline{54.51} \\
\toprule
\multirow{2}{*}{\textbf{Models}} & \multicolumn{9}{c}{\textbf{IWSLT (xx$\rightarrow$en)}} \\
 & \textbf{de} & \textbf{es} & \textbf{fr} & \textbf{it} & \textbf{ko} & \textbf{nl} & \textbf{pt} & \textbf{ru} & \textbf{zh} \\
\midrule
\texttt{Llama-3.3-70B-Instruct} & \underline{84.69} & \underline{88.38} & \underline{88.54} & \underline{88.60} & 69.32 & \textbf{\underline{88.49}} & 81.22 & 77.56 & \underline{73.02} \\
\texttt{Qwen2.5-72B-Instruct} & 83.00 & 87.61 & 87.90 & 87.80 & \underline{71.46} & 85.79 & \textbf{\underline{81.35}} & \underline{77.60} & 71.63 \\
\texttt{GPT-4o} & 82.07 & 83.98 & 82.79 & 85.40 & 65.81 & 79.92 & 77.23 & 74.94 & 68.75 \\
\midrule
\texttt{TowerInstruct-Mistral-7B} & 30.12 & 32.42 & 27.00 & 38.22 & 27.69 & 29.69 & 21.60 & 24.88 & 28.29 \\
\texttt{DocMT-TowerInstruct-Mistral-7B} & \underline{84.28} & \textbf{\underline{90.10}} & \underline{87.97} & \textbf{\underline{91.87}} & \underline{60.02} & \underline{82.90} & \underline{80.16} & \textbf{\underline{79.26}} & \underline{63.16} \\
\midrule
\texttt{EuroLLM-9B-Instruct} & 39.17 & 34.80 & 40.49 & 35.17 & 26.13 & 56.22 & 28.75 & 20.54 & 22.83 \\
\texttt{DocMT-EuroLLM-9B-Instruct} & \underline{77.73} & \underline{83.92} & \underline{81.75} & \underline{86.86} & \underline{69.93} & \underline{79.49} & \underline{80.07} & \underline{75.73} & \underline{68.33} \\
\midrule
\texttt{Qwen2.5-7B-Instruct} & 82.55 & 83.48 & 89.01 & 88.71 & 68.41 & 86.08 & 79.33 & 75.44 & 71.49 \\
\texttt{DocMT-Qwen2.5-7B-Instruct} & \textbf{\underline{85.65}} & \underline{84.49} & \textbf{\underline{91.16}} & \underline{91.72} & \textbf{\underline{74.74}} & \underline{86.82} & \underline{80.88} & \underline{76.26} & \textbf{\underline{85.09}} \\
\bottomrule
\end{tabular}
}
\caption{Evaluation based on \textsc{LTCR} across IWSLT2017 language pairs.}
\label{tab:iwslt-ltcr}
\end{table}

\begin{table}[h]
\centering
\renewcommand{\arraystretch}{1.3} 
\scalebox{0.9}{
\begin{tabular}{lcccccc}
\toprule
\multirow{2}{*}{\textbf{Models}} & \multicolumn{6}{c}{\textbf{WMT23}} \\
 & \textbf{en$\rightarrow$de} & \textbf{en$\rightarrow$ru} & \textbf{en$\rightarrow$zh} & \textbf{de$\rightarrow$en} & \textbf{ru$\rightarrow$en} & \textbf{zh$\rightarrow$en} \\
\midrule
\texttt{TowerInstruct-Mistral-7B} & 83.91 & 85.77 & 86.10 & 85.48 & 83.15 & 80.46 \\
\texttt{DocMT-TowerInstruct-Mistral-7B} & 83.70 & 85.47 & 85.62 & 85.39 & 82.93 & 80.59 \\
\midrule
\texttt{EuroLLM-9B-Instruct} & 84.83 & 85.64 & 86.15 & 85.48 & 83.34 & 80.90 \\
\texttt{DocMT-EuroLLM-9B-Instruct} & 84.55 & 85.62 & 85.61 & 85.31 & 82.72 & 80.61 \\
\midrule
\texttt{Qwen2.5-7B-Instruct} & 78.45 & 76.78 & 84.45 & 81.92 & 78.92 & 78.59 \\
\texttt{DocMT-Qwen2.5-7B-Instruct} & 81.86 & 84.21 & 85.66 & 83.44 & 81.73 & 80.23 \\
\bottomrule
\end{tabular}
}
\caption{Evaluation based on COMET across WMT23 language pairs.}
\label{tab:wmt23-comet-app}
\end{table}

\begin{table}[h]
\centering
\renewcommand{\arraystretch}{1.3} 
\begin{tabular}{lccccc}
\toprule
\multirow{2}{*}{\textbf{Models}} & \multicolumn{5}{c}{\textbf{TICO-19}} \\
 & \textbf{en$\rightarrow$es} & \textbf{en$\rightarrow$fr} & \textbf{en$\rightarrow$pt} & \textbf{en$\rightarrow$ru} & \textbf{en$\rightarrow$zh} \\
\midrule
\texttt{TowerInstruct-Mistral-7B} & 88.60 & 82.03 & 89.51 & 88.31 & 88.85 \\
\texttt{DocMT-TowerInstruct-Mistral-7B} & 88.73 & 81.72 & 89.55 & 88.16 & 87.70 \\
\midrule
\texttt{EuroLLM-9B-Instruct} & 88.61 & 81.82 & 90.01 & 88.48 & 88.74 \\
\texttt{DocMT-EuroLLM-9B-Instruct} & 88.96 & 81.94 & 89.49 & 88.82 & 86.79 \\
\midrule
\texttt{Qwen2.5-7B-Instruct} & 85.51 & 78.17 & 87.56 & 78.77 & 86.79 \\
\texttt{DocMT-Qwen2.5-7B-Instruct} & 88.04 & 81.03 & 88.99 & 87.12 & 87.37 \\
\bottomrule
\end{tabular}
\caption{Evaluation based on COMET across TICO-19 language pairs.}
\label{tab:tico19-comet-app}
\end{table}

\begin{table}[!htbp]
\centering
\renewcommand{\arraystretch}{1.3} 
\scalebox{0.83}{
\begin{tabular}{lccccccccc}
\toprule
\multirow{2}{*}{\textbf{Models}} & \multicolumn{9}{c}{\textbf{FLORES-200 (en$\rightarrow$xx)}} \\
 & \textbf{de} & \textbf{es} & \textbf{fr} & \textbf{it} & \textbf{ko} & \textbf{nl} & \textbf{pt} & \textbf{ru} & \textbf{zh} \\
\midrule
\texttt{TowerInstruct-Mistral-7B} & 88.25 & 87.05 & 88.89 & 89.20 & 90.11 & 88.67 & 89.77 & 90.13 & 88.56 \\
\texttt{DocMT-TowerInstruct-Mistral-7B} & 88.52 & 86.79 & 88.55 & 88.75 & 89.20 & 88.59 & 89.68 & 89.76 & 87.01 \\
\midrule
\texttt{EuroLLM-9B-Instruct} & 88.92 & 87.38 & 89.10 & 89.29 & 89.96 & 88.68 & 90.39 & 90.38 & 88.79 \\
\texttt{DocMT-EuroLLM-9B-Instruct} & 88.71 & 86.94 & 88.74 & 88.94 & 89.81 & 88.59 & 89.39 & 90.26 & 87.73 \\
\midrule
\texttt{Qwen2.5-7B-Instruct} & 83.21 & 84.10 & 84.92 & 83.77 & 77.81 & 80.24 & 87.04 & 79.40 & 87.21 \\
\texttt{DocMT-Qwen2.5-7B-Instruct} & 86.97 & 86.37 & 87.81 & 87.71 & 88.39 & 86.53 & 88.93 & 88.94 & 88.28 \\
\toprule
\multirow{2}{*}{\textbf{Models}} & \multicolumn{9}{c}{\textbf{FLORES-200 (xx$\rightarrow$en)}} \\
 & \textbf{de} & \textbf{es} & \textbf{fr} & \textbf{it} & \textbf{ko} & \textbf{nl} & \textbf{pt} & \textbf{ru} & \textbf{zh} \\
\midrule
\texttt{TowerInstruct-Mistral-7B} & 89.50 & 87.62 & 89.55 & 88.50 & 88.50 & 87.95 & 90.07 & 87.09 & 87.25 \\
\texttt{DocMT-TowerInstruct-Mistral-7B} & 89.47 & 87.24 & 89.44 & 88.20 & 88.30 & 87.67 & 89.78 & 86.84 & 86.95 \\
\midrule
\texttt{EuroLLM-9B-Instruct} & 89.41 & 87.25 & 89.59 & 88.29 & 88.54 & 87.59 & 89.75 & 86.93 & 87.44 \\
\texttt{DocMT-EuroLLM-9B-Instruct} & 89.65 & 87.77 & 89.60 & 88.39 & 88.35 & 87.57 & 90.02 & 87.17 & 87.17 \\
\midrule
\texttt{Qwen2.5-7B-Instruct} & 87.98 & 86.24 & 88.37 & 86.82 & 85.73 & 85.59 & 87.92 & 85.37 & 86.28 \\
\texttt{DocMT-Qwen2.5-7B-Instruct} & 89.25 & 87.38 & 89.20 & 88.04 & 87.74 & 87.41 & 89.58 & 86.77 & 87.16 \\
\bottomrule
\end{tabular}
}
\caption{Evaluation based on COMET across FLORES-200 language pairs.}
\label{tab:flores-comet-app}
\end{table}

\end{document}